%% file: main.tex
\documentclass[preprint,12pt]{elsarticle}

\usepackage{amssymb}
\usepackage{amsmath}

\usepackage{amsmath,amsfonts}
\usepackage{algorithmic}
\usepackage{algorithm}
\usepackage{array}
\usepackage[caption=false,font=normalsize,labelfont=sf,textfont=sf]{subfig}
\usepackage{textcomp}
\usepackage{stfloats}
\usepackage{url}
\usepackage{verbatim}
\usepackage{graphicx}
\usepackage{booktabs}
\usepackage{multirow}
\usepackage{makecell}
\usepackage{subfig}
\usepackage{bm}
\usepackage{color}
\usepackage[normalem]{ulem}

\journal{Information Fusion}

\begin{document}

\begin{frontmatter}



\title{Deep Cut-informed Graph Embedding and Clustering}



\author[a1,a2]{Zhiyuan Ning\fnref{1}}
\author[a1,a2]{Zaitian Wang\fnref{1}}
\author[a1,a2]{Ran Zhang}
\author[a1,a2]{Ping Xu}
\author[a3]{\\Kunpeng Liu}
\author[a4]{Pengyang Wang}
\author[a5]{Wei Ju}
\author[a1,a2]{Pengfei Wang\corref{2}}
\author[a1,a2]{\\Yuanchun Zhou}
\author[a6]{Erik Cambria}
\author[a7]{Chong Chen} 

\affiliation[a1]{organization={Computer Network Information Center, CAS}
}
\affiliation[a2]{organization={University of Chinese Academy of Sciences}
}
\affiliation[a3]{organization={Department of Computer Science, Portland State University}
}
\affiliation[a4]{organization={Department of Computer and Information Science, The State Key Laboratory of Internet of Things for Smart City, University of Macau}
}
\affiliation[a5]{organization={College of Computer Science, Sichuan University}
}
\affiliation[a6]{organization={School of Computer Science and Engineering at Nanyang Technological University}
}
\affiliation[a7]{organization={Future City Lab, Terminus Group}
}

\fntext[1]{Equal contribution. }
\cortext[2]{Corresponding author. }

\input{0_abstract}


\begin{highlights}
\item Investigate graph clustering from a graph cut perspective. 
\item Proposing a novel and non-GNN-based graph clustering paradigm, DCGC. 
\item A cut-informed graph encoding module to fuse graph structure and node attributes. 
\item Self-supervised graph clustering via optimal transport. 
\end{highlights}


\end{frontmatter}



\input{1_introduction}
\input{2_relatedwork}
\input{3_preliminary}
\input{4_method}
\input{5_experiment}
\input{6_conclusion}





\bibliographystyle{elsarticle-num} 
\bibliography{ref}






\end{document}

%% file: 0_abstract.tex
\begin{abstract}
Graph clustering aims to divide the graph into different clusters. The recently emerging deep graph clustering approaches are largely built on graph neural networks (GNN). However, GNN is designed for general graph encoding and there is a common issue of representation collapse in existing GNN-based deep graph clustering algorithms.
We attribute two main reasons for such issues: (i) the inductive bias of GNN models: GNNs tend to generate similar representations for proximal nodes. Since graphs often contain a non-negligible amount of inter-cluster links, the bias results in error message passing and leads to biased clustering; (ii) the clustering guided loss function: most traditional approaches strive to make all samples closer to pre-learned cluster centers, which causes a degenerate solution assigning all data points to a single label thus making all samples similar and less discriminative.  
To address these challenges, we investigate graph clustering from a graph cut perspective and propose an innovative and non-GNN-based {\textbf{D}}eep {\textbf{C}}ut-informed {\textbf{G}}raph embedding and {\textbf{C}}lustering framework, namely \textbf{\textit{DCGC}}. 
This framework includes two modules: (i) cut-informed graph encoding; (ii) self-supervised graph clustering via optimal transport. 
For the encoding module, we derive a cut-informed graph embedding objective to fuse graph structure and attributes by minimizing their joint normalized cut. 
For the clustering module, we utilize the optimal transport theory to obtain the clustering assignments, which can balance the guidance of ``proximity to the pre-learned cluster center”. 
With the above two tailored designs, DCGC is more suitable for the graph clustering task, which can effectively alleviate the problem of representation collapse and achieve better performance.
We conduct extensive experiments to demonstrate that our method is simple but effective compared with benchmarks.
\end{abstract}

%% file: 1_introduction.tex
\begin{figure}[ht]
\centering
\subfloat[GNN-based]{
\includegraphics[width=0.248\textwidth]{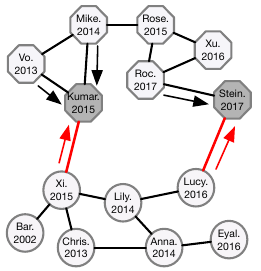}
\label{fig:gcn_noise_a}
}
\subfloat[Cut-informed]{
\includegraphics[width=0.248\textwidth]{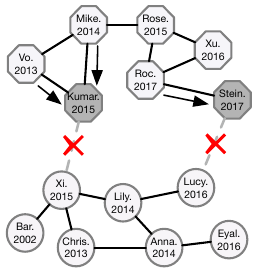}
\label{fig:gcn_noise_b}
}
\caption{The motivation of our deep cut-informed graph embedding and clustering: (a) the GNN-based methods can lead to error message passing when noisy links exist. (b) Our cut-informed approach learns embeddings corresponding to minimal normalized cut.}
\label{fig:gcn_noise}
\end{figure}

\section{Introduction}
Graph clustering is a fundamental task that has attracted intensive attention and achieved great success on many fields, such as co-saliency and community detection~\cite{sun2020network,hu2021multi,xu2024sccdcg}. Beside graph structure, each sample/node is usually accompanied with attributes, such as keywords in citation networks and covariates in social networks. Many researchers have paid attention to combining the graph structure and node attributes to elevate graph clustering performances.

Recently, with the emergence of deep learning~\cite{wang2024comprehensive,zhang2024h2d}, deep graph clustering has become a rapidly evolving research area, which addresses the task of training neural networks to learn representations and partitioning nodes in attribute graphs into distinct groups. 
Promising improvements have been made compared with conventional methods in such an unsupervised mission through utilizing deep neural networks, particularly graph neural networks (GNNs)~\cite{kipf2017semi,veli2018graph,ning2021lightcake}. These models typically embed nodes into the hidden space and then perform clustering algorithms. 
In particular, the work in~\cite{guo2017improved} jointly considered the local structure preservation in deep clustering, optimizing cluster assignments and learning features by integrating the clustering loss and auto-encoder reconstruction loss. 
The work in~\cite{bo2020structural} designed a delivery operator and a dual self-supervised mechanism to combine the auto-encoder representation and the graph convolutional networks (GCNs)~\cite{kipf2017semi} representation. The work in~\cite{tu2021deep} proposed a dynamic cross-modality fusion mechanism and a triplet self-supervised strategy, etc. They all focus on developing sophisticated graph neural networks to fuse graph structure and node attributes to learn a general representation of graph topology (e.g., global, local, neighbor, subgraph) and attributes. 

Despite achieving commendable performance, there is a common issue in existing GNN-based deep graph clustering algorithms. 
Specifically, these algorithms often encounter the problem of representation collapse~\cite{liu2022deep}, wherein nodes belonging to distinct categories are frequently mapped to similar representations during the sample encoding process. 
Consequently, the resulting node representations lack discrimination, leading to limited efficacy in clustering performance.
The issue of representation collapse remains unresolved, which is restricting the performance of deep graph clustering algorithms that rely on GNNs~\cite{liu2022deep,ning2022graph}.

We attribute two main reasons for the representation collapse problem of GNN-based deep graph clustering algorithms: (i) \textbf{\textit{The inductive bias of GNNs}}: 
Many existing GNNs have the inductive bias that linked nodes typically belong to the same class or have similar characteristics~\cite{zhu2020beyond}. 
Such inductive bias makes GNN representations similar for nodes in close proximity~\cite{zhu2021graph}. 
When the GNNs get deeper
, the inductive bias will even lead to the over-smoothing issue (indistinguishable representations of nodes in different classes)~\cite{chen2020measuring}. 
So when faced with graph clustering, an unsupervised learning task without any labeling information, the problem caused by the inductive bias of GNN models will become more severe. 
Besides, the graph structure usually contains a non-negligible amount of inter-cluster links. 
For example, citation networks usually have cross-subject links 
(e.g., Figure \ref{fig:gcn_noise_a}). 
According to the inductive bias, the GNN-based clustering methods may make the node embeddings from different clusters similar, because there may exist error message passing~\cite{gilmer2017neural} through the inter-cluster links (i.e., red lines in Figure \ref{fig:gcn_noise_a}).
Thus, it is urgent to design a new method to reduce the effects of error propagation caused by GNNs' inductive bias.
(ii) \textbf{\textit{The clustering guided loss function}}: 
Most traditional approaches adopt a clustering guided loss function~\cite{xie2016unsupervised} to force the generated sample embeddings to have the minimum distortion against the pre-learned clustering centers. They typically strive to minimize the distance between data representations and pre-learned cluster centers, thereby enhancing cluster cohesion~\cite{wang2019attributed, pan2019learning, bo2020structural, peng2021attention}. 
However, making all samples much closer to cluster centers may cause a degenerate solution that assigns all data points to a single (arbitrary) label, thus making all samples similar and less discriminative~\cite{liu2022deep}. 
Therefore, we need to design a mechanism to balance the guidance of ``proximity to the pre-learned cluster center".

To tackle the above-mentioned issues, we propose a novel and non-GNN-based graph clustering paradigm named \textbf{D}eep \textbf{C}ut-informed \textbf{G}raph embedding and \textbf{C}lustering framework (DCGC). This paradigm includes two modules: 
(i) \textbf{\textit{cut-informed graph encoding}}, which does not use any GNN-based model as graph encoder and designs the core module from a perspective (i.e., graph cut) that is more reasonable and appropriate for the graph clustering task, therefore avoids the problems caused by the inductive bias of GNN models; 
(ii) \textbf{\textit{self-supervised graph clustering via optimal transport}}, which balances the guidance of ``proximity to the pre-learned cluster center” via optimal transport, thus avoiding the degenerate solutions. 
Specifically, for the cut-informed graph encoding module, we show that the normalized cut~\cite{shi2000normalized} minimization can be relaxed into solving $k$-smallest eigenvalues of graph Laplacian, where the eigenvectors are continuous approximations of graph partitions and can be viewed as cut-informed graph embeddings. 
As seen in Figure~\ref{fig:gcn_noise}(b), the cut-informed graph embeddings are much more robust to inter-cluster links. 
Inspired by covariates-assisted spectral clustering~\cite{binkiewicz2017covariate}, we come up with a cut-informed graph embedding objective function to minimize the joint normalized cut of the original graph (i.e., the structure information of graph) and attribute graph (i.e., the attribute information of nodes). 
The embeddings learned by DCGC can better capture the latent clustering structure than existing methods. 
For the self-supervised graph clustering via optimal transport module, we apply an optimal transport~\cite{monge1781memoire} based self-supervised strategy to obtain more stable and balanced clustering assignments. 
Specifically, it transports the original clustering assignments to a new label distribution that is consistent with the cluster size, which can avoid the degenerate solutions of assigning all data points to a single (arbitrary) label. 
Then we force the clustering assignments to be close to the transported distribution by Kullback-Leibler (KL) divergence loss.
With the above two tailored designs, DCGC is more suitable for the task of graph clustering, which can effectively alleviate the problem of representation collapse and achieve better performance.

Our key contributions can be summarized as follows:
\begin{itemize}
\item 
We attribute the problem that the current GNN-based deep graph clustering methods suffer from the phenomenon of representation collapse to two reasons: (i) the inductive bias of GNN models and  (ii) the clustering guided loss function. To tackle these issues, we investigate graph clustering from a graph cut perspective and propose a novel and non-GNN-based graph clustering paradigm, DCGC.
\item 
We propose a cut-informed graph encoding module to fuse the graph structure information and the node attribute information by minimizing their joint normalized cut, thus avoiding the use of GNN models and making our paradigm much more appropriate and suitable for the graph clustering task. 
We show that the normalized cut minimization can be relaxed into solving k-smallest eigenvalues of graph Laplacian, where the eigenvectors are continuous approximations of graph partitions and can be viewed as cut-informed graph embeddings. 
\item 
We propose self-supervised graph clustering via optimal transport to balance the guidance of ``proximity to the pre-learned cluster center” via optimal transport, thus avoiding the degenerate solution of assigning all data points to a single (arbitrary) label. 
\item 
We conduct extensive experiments on 6 challenging real-world graph datasets and the outcomes show our approach can achieve competitive or even superior performance compared with the state-of-the-art deep graph clustering models. We also perform comprehensive ablation experiments to demonstrate that each component of our method is indispensable.
\end{itemize}

%% file: 2_relatedwork.tex
\section{Related Work}
\subsection{Attributed Graph Embedding}
Graph encoding transfers the graph node into vectors~\cite{zhang2024m2mol,zhou2024make,zhang2023htcl}. 
Attributed graph embedding methods assume node attribute information is available and exploit both topological information and attribute features simultaneously~\cite{gao2018deep,zhang2025motif}. ~\cite{yang2015network} proved that DeepWalk can be interpreted as a factorization approach and proposed an extension method to explore node features. DANE~\cite{li2017attributed} deals with the dynamic environment with an incremental matrix factorization approach, and LANE~\cite{huang2017label} incorporates the label information into the optimization process to learn a better embedding. 
Though these methods improve the clustering performance, they fall short in combining the attribute and graph information for graph clustering.


\subsection{Deep Graph Clustering}
Recently, due to the strong representation power of deep neural networks, many deep clustering methods have been proposed and achieved impressive performance~\cite{xie2016unsupervised,guo2017improved}. 
Auto-encoder~\cite{hinton2006reducing} is one of the most commonly used unsupervised deep neural networks, which plays a crucial role in deep clustering. 
DEC~\cite{xie2016unsupervised} is the most popular method which uses the auto-encoder to learn the deep representations by mining divergence between assignment distribution and target distribution. 
To exploit the structural information underlying the data, some GCNs based clustering methods were proposed~\cite{wang2019attributed,bo2020structural,tu2021deep}. 
DAEGC~\cite{wang2019attributed} encodes the topological structure and node contents by introducing the attentional neighbor-wise fusion strategy on the GAE framework.
ARGA~\cite{pan2019learning} adversarially regularized GAE and further improved the clustering performance by introducing an adversarial learning scheme to learn the graph embedding. 
SDCN~\cite{bo2020structural} designs a delivery operator and a dual self-supervised mechanism. 
And DFCN~\cite{tu2021deep} designs a dynamic cross-modality fusion mechanism and a triplet self-supervised strategy. 
MGAI~\cite{liu2024revisiting} leverages modularity maximization as a contrastive pretext task to effectively uncover the underlying information of communities in graphs.
Although these algorithms are well designed for graph data, they are not suitable for unsupervised graph clustering since graph structure often contains a non-negligible amount of inter-cluster links, which lead to considerable error when leveraging structural information by means such as message passing between clusters. 


\subsection{Optimal Transport}
Optimal transport was initially introduced as a solution to the problem of minimizing costs during the simultaneous movement of multiple items~\cite{monge1781memoire}. 
In recent years, its application has garnered significant attention from the machine learning and computer vision community, particularly in the context of comparing distributions represented as feature sets~\cite{peyre2017computational}. 
The remarkable ability of optimal transport to match distributions has led to its utilization in various theoretical and practical tasks, including generative models~\cite{arjovsky2017wasserstein}, structural matching~\cite{chen2019improving}
, and clustering~\cite{laclau2017co}. 
An acknowledged drawback of optimal transportation lies in its computational complexity, necessitating the exploration of approximation techniques~\cite{grauman2004fast}. 
In the pursuit of an efficient solution, a regularization approach influenced by probabilistic theory has been proposed~\cite{cuturi2013sinkhorn}, in conjunction with Sinkhorn's algorithm. 
Alternatively, the problem has been relaxed to achieve a quadratic-time solution by eliminating a constraint~\cite{kusner2015word}. 
Moreover, the incorporation of a kernel method has been introduced as a means to approximate optimal transport~\cite{wu2018word}. 
In this paper, we use optimal transport to get a more appropriate target distribution $P$ which has the constraint that the label distribution must be consistent with the mixing proportions, thus avoiding the degenerate solutions by assigning all data points to a single (arbitrary) label.

%% file: 3_preliminary.tex

\section{Problem Formulation}
Given a graph $\mathcal{G}=(\mathcal{V}, \bm{X}, \mathcal{E}, \bm{A})$, $\mathcal{V}$ is a set of $N$ nodes in the graph, $\bm{X}=[\bm{x}_1, \bm{x}_2, \cdots, \bm{x}_N]^{T}$ is the attribute features of nodes where $\bm{x}_i \in \mathbb{R}^{F}$ and $F$ is the number of attributes, $\mathcal{E} \subseteq \mathcal{V} \times \mathcal{V}$ is the edge set of graph and $\bm{A}\in\{0, 1\}^{N \times N}$ is the adjacency matrix of graph. For $\forall v_{i}, v_{j} \in \mathcal{V}$, $A_{ij}=1$ if there exists an edge between $v_i$ and $v_j$, otherwise, $A_{ij}=0$. Given a vector $\bm{u}$, we write $\|\bm{u}\|_{2}$ as the its Euclidean norm. Given a subset $U\subseteq V$, we write $|U|$ as the number of nodes in $U$. The graph clustering task aims to map each node $v_i \in \mathcal{V}$ to the low-dimensional embedding $\bm{h}_i\in \mathbb{R}^{d}$ based on its original attributes $\bm{x}_i \in \mathbb{R}^{F}$ and the graph structure, and separates the node set $\mathcal{V}$ into $K$ disjoint subsets $\mathcal{V} = \mathcal{V}_{1}\bigcup \cdots \bigcup \mathcal{V}_{K}$ such that each $\mathcal{V}_{k}$ is corresponding to a specific semantic. Assuming $\bm{H}_{N\times d}=[\bm{h}_1, \bm{h}_2, \cdots, \bm{h}_N]^{T}$ is the latent embedding and $\bm{c} = [c_{1},...,c_{N}]$ is the clustering assignment, we will simultaneously learn the embedding $\bm{H}$ and clustering assignments $\bm{c}$.


\begin{table}[htbp]
    \centering
    \vspace{-3mm}
    \footnotesize
    \vspace{-3mm}
    \caption{Notations and Descriptions.}
    \begin{tabular}{lp{110mm}}
    \toprule
    Notations & Descriptions  \\ 
    \midrule

    $\mathcal{G}, \bm{S}$   &   Original graph, and corresponding attribute graph 
 \\ 
    $\mathcal{V}, \bm{X}, \mathcal{E}, \bm{A}$   &   The set of nodes, the attribute features of nodes,  the set of edges, and the adjacency matrix  \\
    $\bm{h}_i, \bm{H}$  &    The low-dimensional embedding of node  $v_i$, and the set of latent embeddings for all nodes\\
    $\bm{I}$ & Identity matrix\\
    $\bm{D}, \bm{D}_{\bm{S}}$  &   Degree matrix of original graph,  and degree matrix of corresponding attribute graph \\
    $L_{\mathcal{G}}, L_{\bm{S}}$ &  The normalized graph Laplacian of $\mathcal{G}$, and the normalized graph Laplacian of $\bm{S}$  \\
    $\text{Tr}$ & The trace of a matrix \\
    $\bm{u}_j$ & Center of cluster $j$ \\
    $q_{ij}$, $Q$ & The real probability distribution of assigning sample $i$ to cluster $j$, and that of all samples \\
    $p_{ij}$, $P$ & The target distribution of sample $i$, and that of all samples \\
    $\hat{P}$ & The optimal transport plan matrix \\ 
    $\bm{\pi}$ & The estimated proportion of points for each cluster (computed during optimization) \\
    $H$ & The entropy function \\
    $\lambda$ &  The smoothness parameter controlling the equilibrium of clusters in Sinkhorn iteration (hyperparameter)\\
    \bottomrule
    \end{tabular}
    
    \vspace{-3mm}
    \label{tab:notation}
\end{table}

%% file: 4_method.tex
\section{Method}

In this section we introduce the \textbf{D}eep \textbf{C}ut-informed \textbf{G}raph embedding and \textbf{C}lustering framework (DCGC). 
Specifically, we propose (i) cut-informed graph encoding, which does not use any GNN-based model as graph encoder and design the core module from a perspective (i.e., graph cut) that is more reasonable and appropriate for the graph clustering task, therefore avoids the problems caused by the inductive bias of GNN models; 
(ii) self-supervised graph clustering via optimal transport, which balances the guidance of ``proximity to the pre-learned cluster center” via optimal transport, thus avoids the degenerated solutions.

\begin{figure*}[htbp]
    \centering
    \includegraphics[width=1\textwidth]{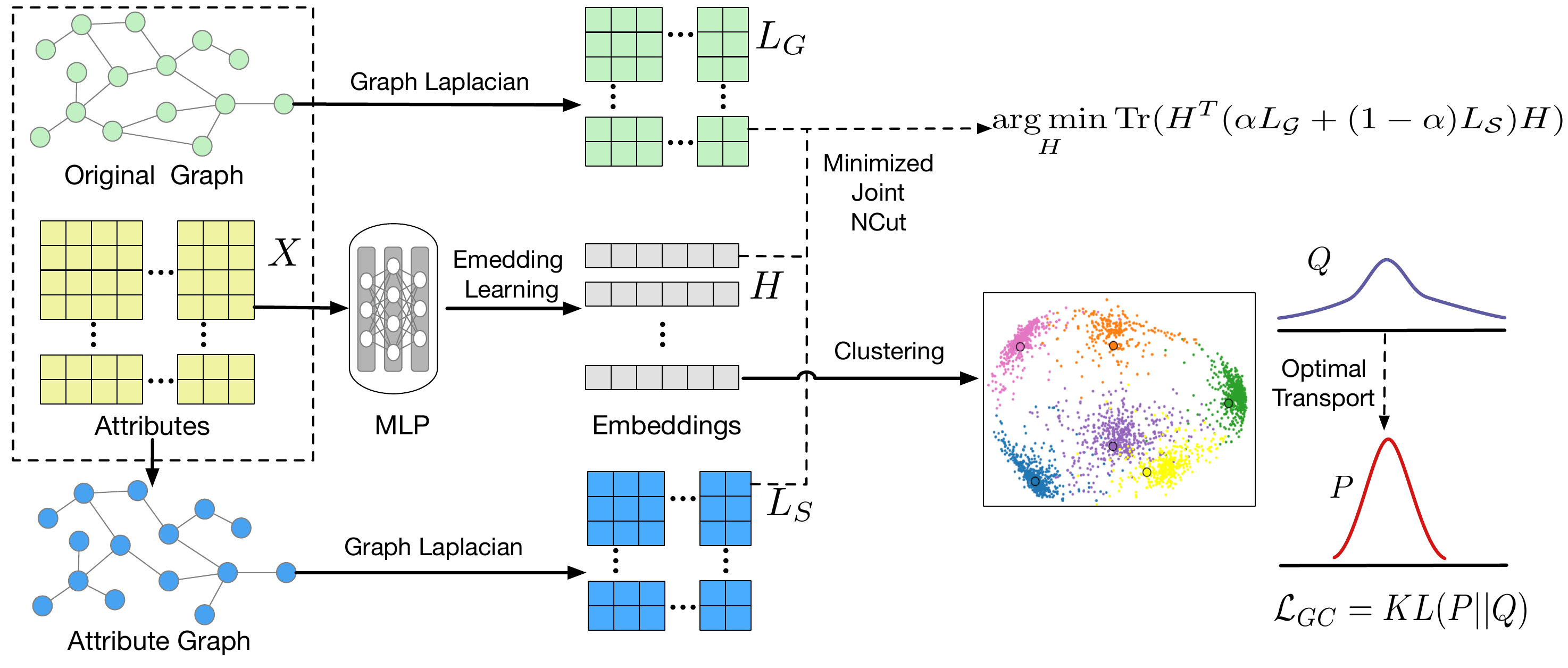}
    \caption{Framework Overview of Cut-informed Graph Clustering: Given original graph and attributes, attribute graph is constructed. Then attributes are encoded via an MLP to obtain the embedding by minimizing the joint normalized cut of original and attribute graphs. The clustering assignments are optimized by a self-supervised strategy with optimal transport target.}
    \label{fig:framework}
\end{figure*}


\subsection{Cut-informed Graph Encoding}
To avoid the problems caused by the inductive bias  of GNNs that they tend to
generate similar representations for nodes in close proximity, instead of using any GNN models as graph encoder, we simply use an MLP to encode the inputted attributes, which can be formulated as:
\begin{equation}
    \bm{H} = \text{MLP}(\bm{X})
\end{equation}
For graph $\mathcal{G}$, denote $\bm{D} = \text{diag}([d_{1},\cdots, d_{N}])$, where $d_{i}$ is the degree of node $i$. The normalized graph Laplacian of $\mathcal{G}$ can be formulated as:
\begin{equation}
\label{equ:lg}
  L_{\mathcal{G}} = \bm{I} - \bm{D}^{-1/2}\bm{A}\bm{D}^{-1/2}.  
\end{equation}

Given two subsets of $\mathcal{V}$ denoted as $\mathcal{V}_{1}$ and $\mathcal{V}_{2}$, we define $\bm{A}(\mathcal{V}_{1}, \mathcal{V}_{2}) = \sum_{i\in \mathcal{V}_{1}, j\in \mathcal{V}_{2}}A_{ij}$. Then for a partition $\mathcal{V} = \mathcal{V}_{1}\bigcup \cdots \bigcup \mathcal{V}_{K}$, its normalized cut~\cite{shi2000normalized} can be defined as:
\begin{equation}
\text{Ncut}(\mathcal{V}_{1},\cdots,\mathcal{V}_{K}) = \frac{1}{2}\sum_{i=1}^{K}\frac{\bm{A}(\mathcal{V}_{i}, \bar{\mathcal{V}}_{i})}{\text{vol}(\mathcal{V}_{i})}
\label{equ:Ncut}
\end{equation}
where $\bar{\mathcal{V}}_{i}$ is the complementary set of $\mathcal{V}_{i}$ and $\text{vol}(\mathcal{V}_{i}) = \bm{A}(\mathcal{V}_{i}, \mathcal{V}_{i}) + \bm{A}(\mathcal{V}_{i}, \bar{\mathcal{V}}_{i})$ is the total edges of $\mathcal{V}_{i}$. According to this criterion, an optimal graph partition should have minimized Ncut. However, it is NP hard to find the optimal graph partition. 
Inspired by covariates-assisted spectral clustering~\cite{binkiewicz2017covariate}, we can alternatively solve it with the following optimization strategy:
\begin{equation}
\begin{aligned}
\label{equ:spectral}
    & \arg\min_{\bm{H}} \text{Tr}(\bm{H}^{T}L_{\mathcal{G}}\bm{H}) \\
    & \text{subject to } \bm{H}^{T}\bm{H} = \bm{I},
\end{aligned}
\end{equation}
which is a relaxation of minimizing the normalized cut. $\bm{H}$ is a soft partition of graph $\mathcal{G}$ in terms of minimizing normalized cut, which implies that $\bm{H}$ can be viewed as the node embeddings for clustering. 
However, the learned embedding $\bm{H}$ by Equation \ref{equ:spectral} only contains the graph structure information and ignores the node attribute information, so it is not optimal for clustering. Inspired by covariates-assisted spectral clustering~\cite{binkiewicz2017covariate}, we construct an attribute graph $\bm{S}$ which is defined as $S_{ij} = \left(1 + \frac{{\bm{x}_i}^{T}\bm{x}_j}{\lVert\bm{x}_i\rVert_2\lVert\bm{x}_j\rVert_2}\right) / 2$ to model the attribute information. It is easy to see $S_{ij}\in [0,1]$. We can also define the graph Laplacian of $\bm{S}$ as: 
\begin{equation}
\label{equ:ls}
L_{\bm{S}} = \bm{I} - \bm{D}_{\bm{S}}^{-1/2}\bm{S}\bm{D}_{\bm{S}}^{-1/2},
\end{equation}
where $\bm{D}_{\bm{S}}$ is a diagonal matrix that corresponding to the row-sums of $\bm{S}$. It is a natural idea to make the embedding $\bm{H}$ contain the structure information of both original graph 
(i.e., the structure information of graph) and attribute graph (i.e., the attribute information of nodes), so we have the follow optimization function:
\begin{equation}
\begin{aligned}
\label{equ:as-spectral}
    & \arg\min_{\bm{H}} \text{Tr}(\bm{H}^{T}(\alpha L_{\mathcal{G}} + (1-\alpha)L_{\bm{S}})\bm{H}) \\
    & \text{subject to } \bm{H}^{T}\bm{H} = \bm{I},
\end{aligned}
\end{equation}
where $\alpha$ is the balance parameter between the original graph and the attribute graph. So the graph encoding loss function can be written as:
\begin{equation}
\label{equ:loss_ge}
    \mathcal{L}_{\text{GE}} = \text{Tr}(\bm{H}^{T}(\alpha L_{\mathcal{G}} + (1-\alpha)L_{\bm{S}})\bm{H}) + \beta \|\bm{H}^{T}\bm{H} - \bm{I}\|_{F},
\end{equation}
where $\beta$ is a tuning parameter and $\|\bm{H}^{T}\bm{H} - \bm{I}\|_{F}$ is the loss to guarantee the orthogonality. 
We chose this weighted sum formulation (Equation~\ref{equ:as-spectral}) due to its balance of effectiveness and simplicity observed across diverse datasets during preliminary exploration, allowing for a clear evaluation of our core contributions.

The learned embedding $\bm{H}$ is suitable for clustering from two viewpoints: (1) The loss function is a relaxation of minimizing the joint normalized cut of the original graph and attribute graph, which means $\bm{H}$ is a soft assignment of the graph partition from the perspective of minimizing normalized cut. 
(2) The trade-off between the original graph and attribute graph makes $\bm{H}$ consistent with both graphs and successfully fuses structure information and attribute information in an elegant and non-GNN-based way.


\subsection{Self-supervised Graph Clustering via Optimal Transport}
To generate clustering assignments, we use a self-supervised strategy proposed by~\cite{xie2016unsupervised}. It uses the Student’s t-distribution~\cite{van2008visualizing} as a kernel to measure the similarity
between embedded point $h_{i}$ and the clustering center $\mu_{j}$:
\begin{equation}
\label{equ:clusassign}
    q_{ij} = \frac{(1 + \|h_{i} - \mu_{j}\|^{2}/\theta)^{-\frac{1 + \theta}{2}}}{\sum_{j^{'}}(1 + \|h_{i} - \mu_{j^{'}}\|^{2}/\theta)^{-\frac{1 + \theta}{2}}}
\end{equation}
where $h_{i}$ is encoded embedding of $x_{i}$ and $\theta$ is the degrees of freedom of the Student’s t-distribution~\cite{van2008visualizing}. $q_{ij}$ can be viewed as the soft assignment of each sample to the clustering centers i.e., the probability of assigning sample $i$ to cluster $j$. Then, we define $Q = [q_{ij}]$ as the distribution of the assignments of all samples.

After obtaining the clustering result distribution Q, we aim to optimize the data representation by learning from the high confidence assignments. Most traditional approaches adopt a clustering guided loss function~\cite{xie2016unsupervised} to force the generated sample embeddings to have the minimum distortion against the pre-learned clustering centers. They typically strive to minimize the distance between data representations and cluster centers, thereby enhancing cluster cohesion. For example, SDCN~\cite{bo2020structural} calculate a target distribution $P = [p_{ij}]$ as follows:
\begin{equation}
    p_{i j}=\frac{q_{i j}^2 / f_j}{\sum_{j^{\prime}} q_{i j^{\prime}}^2 / f_{j^{\prime}}}
\end{equation}
where $f_j = \sum_{i} q_{ij}$ are soft cluster frequencies. In the target distribution $P$, each assignment in $Q$ is squared and normalized so that the assignments have higher confidence~\cite{bo2020structural}. However, making all samples much closer to cluster centers may cause a degenerate solution that assigns all data points to a single (arbitrary) label, and thus make all samples similar and less discriminative~\cite{liu2022deep}. Moreover, it makes the performance of these methods highly dependent on good initial cluster centers, thus leading to manual trial-and-error pre-training.

In order to avoid such problems, we add the constraints that the label distribution must be consistent with the mixing proportions, and we achieve such constraints through optimal transport~\cite{monge1781memoire}. Therefore, we construct the target probability matrix $P$ by solving the following optimal transport strategy:
\begin{equation}
    \begin{aligned}
        \min_{P} \quad & P*(-\text{log}Q) \\
        \mbox{s.t.}\quad & P\in\mathbb{R}_{+}^{N\times C}, \\
         \quad &P\ \bm{1}_C = \bm{1}_N \ \text{ and } \ P^{T}\bm{1}_N = N\bm{\pi}
    \end{aligned}
    \label{equ:opti_1}
\end{equation}
Here we regard the target distribution $P$ as the transport plan matrix~\cite{monge1781memoire} in the optimal transport theory, and the $-\text{log}Q$ as the cost matrix~\cite{monge1781memoire} in the optimal transport theory. And we restrict $ P^{T}\bm{1}_N = N\bm{\pi}$, $\bm{\pi}$ is the proportion of points for each cluster, which can be estimated by the intermediate clustering result, thus we successfully add the constraints that the result cluster distribution must be consistent with the mixing proportions. 

As directly optimizing the above objective is always time-consuming, we apply the Sinkhorn distance~\cite{sinkhorn1967diagonal} to use an entropic constraint for fast optimization. The optimization problem with a Lagrange multiplier of the entropy constraint is as follows:
\begin{equation}
    \begin{aligned}
        \min_{P} \quad & P*(-\text{log}Q)-\frac{1}{\lambda}\text{H}(P) \\
        \mbox{s.t.}\quad & P\in\mathbb{R}_{+}^{N\times C}, \\
         \quad &P\ \bm{1}_C = \bm{1}_N \ \text{ and } \ P^{T}\bm{1}_N = N\bm{\pi}
    \end{aligned}
    \label{equ:opti_2}  
\end{equation}
where $H$ is the entropy function and $\lambda$ is the smoothness parameter controlling the equilibrium of clusters. The  existence and unicity of $P$ are guaranteed, and can be efficiently solved by Sinkhorn’s~\cite{sinkhorn1967diagonal} fixed point iteration:
\begin{equation}
\label{equ:sdcn}
    \hat{P}^{(t)}=\operatorname{diag}\left(\boldsymbol{u}^{(t)}\right) Q^{\lambda} \operatorname{diag}\left(\boldsymbol{v}^{(t)}\right)
\end{equation}
where $t$ denotes the iteration and in each iteration: 
$\boldsymbol{u}^{(t)}= \bm{1}_N / (Q^{\lambda} \boldsymbol{v}^{(t-1)})$ 
and
$\boldsymbol{v}^{(t)}= N\bm{\pi} / (Q^{\lambda} \boldsymbol{u}^{(t)} )$, 
with the initiation $\boldsymbol{v}^{(0)}=\bm{1}_N$. 
After the fixed point iteration, we can get the optimal transport plan matrix $\hat{P}$, i.e., the solution of (\ref{equ:opti_1}). 

Then during the training process, we fix $\hat{P}$ and make $Q$ be consistent with $\hat{P}$, so the graph clustering loss function can be formulated as:
\begin{equation}
\label{equ:loss_gc}
    \mathcal{L}_{GC} = \text{KL}(\hat{P}||Q) = \sum_i \sum_j \hat{p}_{i j} \log \frac{\hat{p}_{i j}}{q_{i j}}.
\end{equation}

Thus, the overall loss function is
\begin{equation}
    \mathcal{L} = \mathcal{L}_{GE} + \gamma \mathcal{L}_{GC},
    \label{equ:total_loss}
\end{equation}
where $\gamma$ is a tuning parameter to balance the two losses, $\mathcal{L}_{GE}$ and $\mathcal{L}_{GC}$.

\vspace{-3mm}

\subsection{Overall algorithm}
\label{sec:algo}
We first train the model with cut-informed graph encoding loss. After that, we apply K-means to the embeddings to obtain the initial centroids and cluster size. Then we train the model with both graph encoding loss and graph clustering loss. Then clustering results are obtained at the last epoch. The details are summarized in the Algorithm \ref{alg:DCGC}.
\begin{algorithm}
    \caption{Deep Cut-informed Graph Clustering}
    \begin{algorithmic}[1]
        \REQUIRE Attribute feature matrix $\bm{X}$;\\
                \quad \ \ Graph adjacent matrix $\bm{A}$;\\
                \quad \ \ Number of clusters $K$;\\
                \quad \ \ The number of iteration $M$.
        \ENSURE Clustering results $\bm{c}=\{c_i\}_{i=1}^{N}$.\\
        \STATE Calculate the graph Laplacian of original graph and attribute graph by Equation \ref{equ:lg} and \ref{equ:ls}.
        \FOR{iteration from 1 to $M$}
            \STATE Generate node embedding $\bm{H}^{(t)}$;
            \STATE Compute the graph encoding loss by Equation \ref{equ:loss_ge};
            \STATE Compute clustering assignments by Equation \ref{equ:clusassign} and target distribution by optimal transport strategy, i.e. \ref{equ:opti_1};
            \STATE Compute the graph clustering loss by Equation \ref{equ:loss_gc};
            \STATE Update the parameters by minimizing Equation \ref{equ:total_loss}
        \ENDFOR
        \STATE Return the clustering results $\bm{c}^{(M)}$.
    \end{algorithmic}
    \label{alg:DCGC}
\end{algorithm}

%% file: 5_experiment.tex
\vspace{-6mm}

\section{Experiment}
In this section, we conduct a series of experiments to validate the efficacy of our proposed DCGC framework across six datasets. The results demonstrate the superiority of our method when compared to twelve state-of-the-art approaches.
We conducted sufficient ablation experiments to verify the effectiveness of each module of our framework.

\begin{table*}
    \centering
    \caption{The statistics of the datasets.}
    \begin{tabular}{lcccc}
    \toprule
    Dataset & Network Type & Nodes & Classes & Dimension \\ 
    \midrule
    Cora & Citation & 2708 & 7 & 1433 \\
    ACM & Paper & 3025 & 3 & 1870 \\
    DBLP & Author & 4057 & 4 & 334 \\
    Citeseer & Citation & 3327 & 6 & 3703 \\
    Amazon & Item & 7650 & 8 & 745 \\
    PubMed & Citation & 19717 & 3 & 500 \\
    \bottomrule
    \end{tabular}
    \label{tab:data}
\end{table*}

\vspace{-3mm}

\subsection{Datasets}
We evaluate the proposed method on six public benchmark datasets including multiple types of graphs~\cite{yue2022survey}. 
The statistical information of these datasets is provided in Table \ref{tab:data} and more detailed descriptions are as follows:
\textbf{Cora} is a citation network that comprises scientific publications categorized into 7 distinct classes. Each publication is represented by a binary word vector. 
\textbf{ACM} is a paper dataset containing papers categorized into 3 classes based on their research area. An edge indicates co-authorship of two papers. The paper features are represented by the bag-of-words of the keywords. 
\textbf{DBLP} is a co-authorship network where authors are categorized based on the domain of conferences to which they have contributed. The author features are represented by the bag-of-words of their keywords.
\textbf{Citeseer} is a citation network where each publication within the dataset is represented by a binary word vector, with a value of 0 or 1, denoting the absence or presence. 
\textbf{Amazon} is a co-purchase graph of products belonging to different categories. The edges indicate two products are frequently purchased together. The features represent the product reviews encoded using a bag-of-words approach.
\textbf{PubMed} is a citation network consisting of scientific publications from the PubMed database. Each publication is described by a TF/IDF weighted word vector.

\noindent


\subsection{Baselines}
We compare the performance of our proposed method with twelve baseline methods: 
\textbf{K-means}~\cite{hartigan1979algorithm} is a classical clustering method based on the raw data.
\textbf{AE}~\cite{hinton2006reducing} performs K-means on the representations learned by auto-encoder.
\textbf{DEC}~\cite{xie2016unsupervised} employs a clustering loss to supervise the process of clustering with the auto-encoder backbone.
\textbf{GAE}~\cite{kipf2016variational} is a framework for unsupervised learning on graph-structured data based on the variational auto-encoder.
\textbf{DAEGC}~\cite{wang2019attributed} uses an attention network to learn the node representations and employs a clustering loss to supervise the process of graph clustering.
\textbf{ARGA}~\cite{pan2019learning} applies adversarial training to enforce latent code to match prior distribution.
\textbf{MVGRL}~\cite{hassani2020contrastive} learns representations by contrasting encodings from first-order neighbors and a graph diffusion.
\textbf{SDCN}~\cite{bo2020structural} is representative of hybrid methods which take advantage of both AE and GCN modules for clustering.
\textbf{AGCN}~\cite{peng2021attention} utilizes the attention-based method by considering the dynamic fusion strategy and the multi-scale features fusion.
\textbf{DFCN}~\cite{tu2021deep} designs a dynamic cross-modality fusion mechanism and a triplet self-supervised strategy.
\textbf{DRCN}~\cite{liu2022deep} reduces information correlation in a dual manner to address the issue of representation collapse.
\textbf{GC-SEE}~\cite{ding2023graph} simultaneously considers and exploits different types of structure information to extract more comprehensive features. 
\textbf{CDNMF}~\cite{li2024contrastive} uses nonnegative matrix factorization, combined with contrastive learning between network topology and node attribute. 
\textbf{MGAI}~\cite{liu2024revisiting} leverages modularity maximization as a contrastive pretext task to effectively uncover the underlying information of communities in graphs. 


\subsection{Metrics}
To show the effectiveness of the proposed method,
we employ four popular metrics~\cite{xia2014robust}. For each metric, the larger value implies a better clustering result. 
The best map between cluster ID and class ID is found by using the Kuhn-Munkres algorithm~\cite{lovasz1986matching}.
The four specific evaluation metrics are as follows: 
\textbf{ACC}: Accuracy shows the quality between the predicted labels and the true labels. It can be computed by $ACC=\frac{\sum_{n=1}^{N}\mathcal{I}_n}{N}$, where $\mathcal{I}_{n}$ is an indicator function, $\mathcal{I}_n=1$ when the predicted label and the true label are the same, and $\mathcal{I}_n=0$ otherwise. 
\textbf{NMI}: Normalized Mutual Information, a symmetric index computing the similarity between two clustering solutions based on the confusion matrix (also referred to as the contingency matrix). 
\textbf{ARI}: Adjusted Rand Index, ARI shows the ratio of the number of node pairs similarly classified in both solutions, divided by the total number of pairs. It compares two clusterings with the number of cluster membership agreements and disarrangements. 
\textbf{F1}: Macro F1 score can combine the precision and recall into a single metric by taking their harmonic mean with equation $F1=\frac{2*Precision*Recall}{Precision+Recall}$, where $Precision=\frac{TP}{TP+FP}$ and $Recall=\frac{TP}{TP+FN}$. 
ACC and F1 are computed after achieving the best map between the class ID and the cluster ID with the Kuhn-Munkres algorithm.

\begin{table*}[htbp]
\centering
\caption{Clustering results on six benchmark datasets (mean$\pm$std). The bold values represent the best results.  
}
\scriptsize
\label{tab:result}
\renewcommand{\arraystretch}{0.95} 
\setlength{\tabcolsep}{0.01mm}{
\begin{tabular}{@{}llcccclcccc@{}}
\toprule
                &  & \multicolumn{4}{c}{\textbf{Cora}}     
                &  & \multicolumn{4}{c}{\textbf{ACM}}     \\ \cmidrule(lr){3-6} \cmidrule(lr){8-11} 
\textbf{Method} &  & ACC               & NMI               & ARI               & F1     
                &  & ACC               & NMI               & ARI               & F1    \\ \midrule
K-Means         &  & $33.80_{\pm2.71}$ & $14.98_{\pm3.43}$ & $08.60_{\pm1.95}$ & $30.26_{\pm4.46}$ 
                &  & $67.31_{\pm0.71}$ & $32.44_{\pm0.46}$ & $30.60_{\pm0.69}$ & $67.57_{\pm0.74}$ \\
AE              &  & $49.38_{\pm0.91}$ & $25.65_{\pm0.65}$ & $21.63_{\pm0.58}$ & $43.71_{\pm1.05}$ 
                &  & $81.83_{\pm0.08}$ & $49.30_{\pm0.16}$ & $54.64_{\pm0.16}$ & $82.01_{\pm0.08}$ \\
DEC             &  & $46.50_{\pm0.26}$ & $23.54_{\pm0.34}$ & $15.13_{\pm0.42}$ & $39.23_{\pm0.17}$ 
                &  & $84.33_{\pm0.76}$ & $54.54_{\pm1.51}$ & $60.64_{\pm1.87}$ & $84.51_{\pm0.74}$ \\
GAE             &  & $43.38_{\pm2.11}$ & $28.78_{\pm2.97}$ & $16.43_{\pm1.65}$ & $33.48_{\pm3.05}$ 
                &  & $84.52_{\pm1.44}$ & $55.38_{\pm1.92}$ & $59.46_{\pm3.10}$ & $57.36_{\pm0.82}$ \\
DAEGC           &  & $70.43_{\pm0.36}$ & $52.89_{\pm0.69}$ & $49.63_{\pm0.43}$ & $68.27_{\pm0.57}$ 
                &  & $86.94_{\pm2.83}$ & $56.18_{\pm4.15}$ & $59.35_{\pm3.89}$ & $87.07_{\pm2.79}$ \\
ARGA            &  & $71.04_{\pm0.25}$ & $51.06_{\pm0.52}$ & $47.71_{\pm0.33}$ & $69.27_{\pm0.39}$ 
                &  & $86.29_{\pm0.36}$ & $56.21_{\pm0.82}$ & $63.37_{\pm0.86}$ & $58.23_{\pm0.31}$ \\
MVGRL           &  & $70.47_{\pm3.70}$ & $55.57_{\pm1.54}$ & $48.70_{\pm3.94}$ & $67.15_{\pm1.86}$ 
                &  & $86.73_{\pm0.76}$ & $60.87_{\pm1.40}$ & $65.07_{\pm1.76}$ & $63.71_{\pm0.39}$ \\
SDCN            &  & $35.60_{\pm2.83}$ & $14.28_{\pm1.91}$ & $07.78_{\pm1.24}$ & $24.37_{\pm1.04}$ 
                &  & $90.45_{\pm0.18}$ & $68.31_{\pm0.25}$ & $73.91_{\pm0.40}$ & $90.42_{\pm0.19}$ \\
AGCN            &  & $47.67_{\pm1.25}$ & $26.59_{\pm1.31}$ & $18.36_{\pm1.36}$ & $33.78_{\pm1.49}$ 
                &  & $90.59_{\pm0.15}$ & $68.38_{\pm0.45}$ & $74.20_{\pm0.38}$ & $90.58_{\pm0.17}$ \\
DFCN            &  & $36.33_{\pm0.49}$ & $19.36_{\pm0.87}$ & $04.67_{\pm2.10}$ & $26.16_{\pm0.50}$ 
                &  & $90.84_{\pm0.15}$ & $69.39_{\pm0.36}$ & $74.93_{\pm0.37}$ & $90.78_{\pm0.16}$ \\
DCRN            &  & $72.67_{\pm0.99}$ & $56.09_{\pm1.17}$ & $50.22_{\pm2.09}$ & $67.59_{\pm0.86}$ 
                &  & $91.93_{\pm0.20}$ & $71.59_{\pm0.61}$ & $77.56_{\pm0.61}$ & $91.94_{\pm0.20}$ \\
GC-SEE          &  & $73.58_{\pm0.74}$ & $53.02_{\pm0.41}$ & $51.22_{\pm0.88}$ & $71.48_{\pm0.79}$ 
                &  & $91.67_{\pm0.10}$ & $70.83_{\pm0.25}$ & $76.89_{\pm0.24}$ & $91.66_{\pm0.09}$ \\
CDNMF           &  & $58.59_{\pm3.44}$ & $38.74_{\pm2.57}$ & $35.72_{\pm3.51}$ & $52.69_{\pm5.32}$ 
                &  & $62.94_{\pm4.67}$ & $20.91_{\pm1.77}$ & $21.21_{\pm4.58}$ & $64.56_{\pm3.48}$ \\
MGAI            &  & $73.52_{\pm0.62}$ & \bm{$57.05_{\pm0.91}$} & $52.24_{\pm1.26}$ & $72.13_{\pm0.50}$ 
                &  & $90.00_{\pm0.36}$ & $66.46_{\pm0.77}$ & $72.19_{\pm1.01}$ & $90.05_{\pm0.35}$ \\
\textbf{DCGC}   &  & \bm{$75.41_{\pm1.69}$} & $56.55_{\pm2.00}$ & \bm{$53.61_{\pm1.94}$} & \bm{$72.15_{\pm3.12}$} 
                &  & \bm{$92.50_{\pm0.32}$} & \bm{$72.98_{\pm0.60}$} & \bm{$78.87_{\pm0.63}$} & \bm{$92.51_{\pm0.23}$} \\
\bottomrule 
\toprule
                &  & \multicolumn{4}{c}{\textbf{DBLP}}     
                &  & \multicolumn{4}{c}{\textbf{PubMed}}   \\ \cmidrule(lr){3-6} \cmidrule(lr){8-11} 
\textbf{Method} &  & ACC               & NMI               & ARI               & F1     
                &  & ACC               & NMI               & ARI               & F1    \\ \midrule
K-Means         &  & $38.65_{\pm0.65}$ & $11.45_{\pm0.38}$ & $06.97_{\pm0.39}$ & $31.92_{\pm0.27}$ 
                &  & $59.83_{\pm0.01}$ & $31.05_{\pm0.02}$ & $28.10_{\pm0.01}$ & $58.88_{\pm0.01}$  \\
AE              &  & $51.43_{\pm0.35}$ & $25.40_{\pm0.16}$ & $12.21_{\pm0.43}$ & $52.53_{\pm0.36}$ 
                &  & $63.07_{\pm0.31}$ & $26.32_{\pm0.57}$ & $23.86_{\pm0.57}$ & $64.01_{\pm0.29}$  \\
DEC             &  & $58.16_{\pm0.56}$ & $29.51_{\pm0.28}$ & $23.92_{\pm0.39}$ & $59.38_{\pm0.51}$  
                &  & $60.14_{\pm0.09}$ & $22.44_{\pm0.14}$ & $19.55_{\pm0.13}$ & $61.49_{\pm0.10}$  \\ 
GAE             &  & $61.21_{\pm1.22}$ & $30.80_{\pm0.91}$ & $22.02_{\pm1.40}$ & $61.41_{\pm2.23}$  
                &  & $62.09_{\pm0.81}$ & $23.84_{\pm3.54}$ & $20.62_{\pm1.39}$ & $68.08_{\pm1.76}$  \\ 
DAEGC           &  & $62.05_{\pm0.48}$ & $32.49_{\pm0.45}$ & $21.03_{\pm0.52}$ & $61.75_{\pm0.67}$  
                &  & $68.73_{\pm0.03}$ & $28.26_{\pm0.03}$ & $29.84_{\pm0.04}$ & $68.23_{\pm0.02}$  \\ 
ARGA            &  & $64.83_{\pm0.59}$ & $29.42_{\pm0.92}$ & $27.99_{\pm0.91}$ & $64.97_{\pm0.66}$  
                &  & $65.26_{\pm0.12}$ & $24.80_{\pm0.17}$ & $24.35_{\pm0.17}$ & $65.69_{\pm0.13}$  \\ 
MVGRL           &  & $42.73_{\pm1.02}$ & $15.41_{\pm0.63}$ & $08.22_{\pm0.50}$ & $40.52_{\pm1.51}$  
                &  & $67.01_{\pm0.52}$ & $31.59_{\pm1.45}$ & $29.42_{\pm1.06}$ & $67.07_{\pm0.36}$  \\ 
SDCN            &  & $68.05_{\pm1.81}$ & $39.50_{\pm1.34}$ & $39.15_{\pm2.01}$ & $67.71_{\pm1.51}$  
                &  & $64.20_{\pm1.30}$ & $22.87_{\pm2.04}$ & $22.30_{\pm2.07}$ & $65.01_{\pm1.21}$  \\ 
AGCN            &  & $73.26_{\pm0.37}$ & $39.68_{\pm0.42}$ & $42.49_{\pm0.31}$ & $72.80_{\pm0.56}$  
                &  & $64.70_{\pm0.90}$ & $26.20_{\pm2.30}$ & $24.10_{\pm2.00}$ & $52.70_{\pm1.90}$  \\ 
DFCN            &  & $76.02_{\pm0.77}$ & $43.65_{\pm0.01}$ & $46.95_{\pm1.51}$ & $75.74_{\pm0.75}$  
                &  & $68.89_{\pm0.07}$ & $31.43_{\pm0.13}$ & $30.64_{\pm0.11}$ & $68.10_{\pm0.07}$  \\ 
DCRN            &  & \bm{$79.66_{\pm0.25}$} & \bm{$48.95_{\pm0.44}$} & \bm{$53.60_{\pm0.46}$} & \bm{$79.28_{\pm0.26}$} 
                &  &        OOM        &        OOM        &        OOM        &        OOM         \\ 
GC-SEE          &  & $79.23_{\pm0.96}$ & $48.04_{\pm1.46}$ & $53.51_{\pm1.82}$ & $78.55_{\pm0.99}$ 
                &  & $67.78_{\pm0.38}$ & $25.62_{\pm1.22}$ & $25.13_{\pm0.86}$ & $67.72_{\pm0.34}$  \\ 
CDNMF           &  & $39.48_{\pm3.18}$ & $07.22_{\pm2.19}$ & $05.29_{\pm2.45}$ & $37.91_{\pm3.16}$ 
                &  & $64.71_{\pm3.04}$ & $21.35_{\pm3.09}$ & $25.47_{\pm3.75}$ & $55.10_{\pm2.52}$ \\
MGAI            &  & $77.38_{\pm0.87}$ & $48.17_{\pm0.83}$ & $49.96_{\pm1.52}$ & $77.04_{\pm0.83}$ 
                &  & $62.79_{\pm1.00}$ & $22.77_{\pm1.75}$ & $18.62_{\pm1.71}$ & $62.52_{\pm0.83}$ \\
\textbf{DCGC}   &  & $78.05_{\pm0.56}$ & $46.46_{\pm1.22}$ & $50.68_{\pm0.94}$ & $77.51_{\pm0.66}$  
                &  & \bm{$68.91_{\pm0.43}$} & \bm{$32.53_{\pm1.65}$} & \bm{$30.81_{\pm0.85}$} & \bm{$68.74_{\pm0.62}$}  \\ 
\bottomrule
\toprule
                &  & \multicolumn{4}{c}{\textbf{Citeseer}} 
                &  & \multicolumn{4}{c}{\textbf{Amazon}}  \\ \cmidrule(lr){3-6} \cmidrule(lr){8-11} 
\textbf{Method} &  & ACC               & NMI               & ARI               & F1     
                &  & ACC               & NMI               & ARI               & F1    \\ \midrule
K-Means         &  & $39.32_{\pm3.17}$ & $16.94_{\pm3.22}$ & $13.43_{\pm3.02}$ & $36.08_{\pm3.53}$ 
                &  & $43.24_{\pm4.37}$ & $30.74_{\pm4.33}$ & $17.78_{\pm2.82}$ & $30.34_{\pm7.45}$ \\
AE              &  & $57.08_{\pm0.13}$ & $27.64_{\pm0.08}$ & $29.31_{\pm0.14}$ & $53.80_{\pm0.11}$   
                &  & $59.72_{\pm3.87}$ & $51.89_{\pm3.70}$ & $40.47_{\pm3.06}$ & $47.76_{\pm6.04}$ \\
DEC             &  & $55.89_{\pm0.20}$ & $28.34_{\pm0.30}$ & $28.12_{\pm0.36}$ & $52.62_{\pm0.17}$   
                &  & $59.84_{\pm0.24}$ & $54.67_{\pm0.30}$ & $42.21_{\pm0.25}$ & $47.72_{\pm2.87}$ \\  
GAE             &  & $61.35_{\pm0.80}$ & $34.63_{\pm0.65}$ & $33.55_{\pm1.18}$ & $57.36_{\pm0.82}$ 
                &  & $71.57_{\pm2.48}$ & $62.13_{\pm2.79}$ & $20.62_{\pm1.39}$ & $68.08_{\pm1.76}$ \\  
DAEGC           &  & $64.54_{\pm1.39}$ & $36.41_{\pm0.86}$ & $37.78_{\pm1.24}$ & $62.20_{\pm1.32}$   
                &  & $71.56_{\pm3.34}$ & $60.68_{\pm2.58}$ & $52.05_{\pm3.76}$ & $67.55_{\pm3.39}$ \\  
ARGA            &  & $61.07_{\pm0.49}$ & $34.40_{\pm0.71}$ & $34.32_{\pm0.70}$ & $58.23_{\pm0.31}$   
                &  & $69.28_{\pm2.30}$ & $58.36_{\pm2.76}$ & $44.18_{\pm4.41}$ & $64.30_{\pm1.95}$ \\ 
MVGRL           &  & $68.66_{\pm0.36}$ & $43.66_{\pm0.40}$ & $44.27_{\pm0.73}$ & $63.71_{\pm0.39}$   
                &  & $53.44_{\pm0.81}$ & $36.89_{\pm1.31}$ & $29.42_{\pm1.06}$ & $39.65_{\pm2.39}$ \\ 
SDCN            &  & $65.96_{\pm0.31}$ & $38.71_{\pm0.32}$ & $40.17_{\pm0.43}$ & $63.62_{\pm0.24}$   
                &  & $75.51_{\pm1.92}$ & $63.26_{\pm2.05}$ & $54.95_{\pm2.23}$ & $69.44_{\pm1.34}$ \\ 
AGCN            &  & $68.79_{\pm0.23}$ & $41.54_{\pm0.30}$ & $43.79_{\pm0.31}$ & $62.37_{\pm0.20}$   
                &  & $76.80_{\pm0.40}$ & $63.17_{\pm0.72}$ & $55.67_{\pm0.84}$ & $68.32_{\pm0.62}$ \\ 
DFCN            &  & $69.54_{\pm0.15}$ & $43.93_{\pm0.22}$ & $45.45_{\pm0.26}$ & $64.27_{\pm0.20}$   
                &  & $79.13_{\pm0.90}$ & $71.12_{\pm0.98}$ & $62.41_{\pm1.58}$ & $72.92_{\pm0.81}$ \\  
DCRN            &  & $70.86_{\pm0.18}$ & $45.86_{\pm0.35}$ & $47.64_{\pm0.30}$ & $65.83_{\pm0.21}$   
                &  & $79.94_{\pm0.13}$ & \bm{$73.70_{\pm0.24}$} & $63.69_{\pm0.20}$ & $73.82_{\pm0.12}$ \\ 
GC-SEE          &  & $70.90_{\pm0.56}$ & $44.00_{\pm0.64}$ & $46.47_{\pm0.76}$ & $63.12_{\pm0.66}$   
                &  & $77.34_{\pm0.80}$ & $64.15_{\pm0.68}$ & $56.76_{\pm1.24}$ & $74.56_{\pm0.77}$ \\ 
CDNMF           &  & $46.27_{\pm1.74}$ & $24.67_{\pm1.88}$ & $21.48_{\pm2.46}$ & $41.79_{\pm1.79}$ 
                &  & $65.20_{\pm0.19}$ & $52.85_{\pm0.11}$ & $43.90_{\pm0.08}$ & $56.08_{\pm0.09}$ \\
MGAI            &  & $70.13_{\pm0.35}$ & $44.41_{\pm0.55}$ & $46.11_{\pm0.42}$ & $65.11_{\pm0.34}$ 
                &  & $78.86_{\pm0.66}$ & $69.86_{\pm0.87}$ & $61.98_{\pm0.97}$ & $72.43_{\pm0.40}$ \\
\textbf{DCGC}   &  & \bm{$71.36_{\pm0.30}$} & \bm{$46.25_{\pm0.48}$} & \bm{$47.75_{\pm0.55}$} & \bm{$66.04_{\pm0.94}$}   
                &  & \bm{$80.17_{\pm0.13}$} & $71.77_{\pm0.37}$ & \bm{$66.70_{\pm0.38}$} & \bm{$75.59_{\pm0.34}$} \\ 
\bottomrule

\end{tabular}
}
\end{table*}


\subsection{Implementation Details and Hyper-parameters Setting}
We implement our proposed model\footnote{The code is available at \url{https://github.com/CNICDS/DCGC}} based on PyTorch 2.0.1 and run experiments on one A100 40GB GPU with CUDA version 12.4. 
A fixed MLP architecture was utilized as the encoder across all datasets to ensure fair comparison and reproducibility. 
The embedding size $d$ is fixed to $16$ for all datasets. 
We optimize DCGC with Adam optimizer with learning rate $0.005$ for ACM and Citeseer, $0.001$ for DBLP, Amazon, and Cora, and $0.01$ for PubMed; 
We set the weight decay $0.005$ for ACM, Citeseer, DBLP, PubMed, and Amazon, and $0.0005$ for Cora.
The terms in the loss functions, balance parameter $\alpha$ is set as $0.5$ for ACM, $0.4$ for Citesser, $0.8$ for DBLP, and $0.9$ for Amazon, PubMed and Cora;
tuning parameter $\beta$ is set as $2$ for PubMed, $4$ for ACM and Amazon, $1.5$ for Citesser, $5$ for DBLP, and $20$ for Cora;
tuning parameter $\gamma$ is set as $0.4$ for ACM, $0.25$ for Citesser, $1$ for DBLP, $0.5$ for Amazon and PubMed, and $2$ for Cora;
the smoothness parameter $\lambda$ is set as $20$ for Amazon and PubMed, and $5$ for the other four datasets;
the degrees of freedom of the Student’s t-distribution $\theta$ is $1$ for all datasets;
All hyper-parameters are achieved by the grid search method. 
For baseline methods compared in Table~\ref{tab:result}, we primarily utilized the hyperparameter settings reported in their original papers or the default configurations provided in their official implementations. 
For both stages in Section~\ref{sec:algo}, we train the model for 100 epochs. (Taking 4 minutes on average.)
Results are the average over 10 runs with random seeds, and we report the mean values and the corresponding standard deviations. 


\subsection{Overall Clustering Performance}

The overall results of the clustering performance of all compared methods on six benchmark datasets are shown in Table \ref{tab:result}. From these results, we can have the following observations:
DCGC achieves the best results on five out of six datasets. 
Only on DBLP is our method sub-optimal. 
A possible explanation is that DBLP is not fully connected, and based on spectral embedding, DCGC may not work as expected.
Compared with deep learning-based methods, our proposed DCGC avoids error message passing in GCNs or degenerated clustering results by minimizing the normalized cut and using the optimal transport theorem to guarantee the clustering process. Thus, DCGC can achieve better performance than SOTA baselines.
In particular, compared with the two SOTA results of the baselines (DCRN and GC-SEE), our approach achieves a significant improvement of $0.99\%$ on ACC, $3.79\%$ on NMI, $1.93\%$ on ARI, and $1.46\%$ on F1 score, averagely.


\subsection{Ablation Study}
We conduct ablation studies to evaluate the contributions of different components in our method.
\paragraph{Effect of graph information.} 
To demonstrate both original graph and attribute graph are important, we conduct ablation studies on six datasets with four metrics. 
\textit{DCGC - w/o OA} denotes the model without cut-informed graph encoding. 
\textit{DCGC - w/o O}, and \textit{DCGC - w/o A} denote the models without the original graph and the attribute graph, respectively. 
Table \ref{tab:ablation_feature} shows that each graph improves the performance compared with the baseline. 
The results of \textit{DCGC - w/o A} are better than those of \textit{DCGC - w/o O} for all data sets and all metrics, indicating that the gain from introducing structural information of the graph is greater than that from introducing attribute information of the graph in the graph clustering task.
And the final model DCGC with both the original graph and attribute graph performs especially better than all these baselines, which implies that the two graphs can interact to obtain better embeddings for the graph clustering task.

\begin{table}[htbp!]
    \small
    \centering
    \caption{Ablation study on original graph and attribute graph (mean$\pm$std). 
    }
    \label{tab:ablation_result}
    \setlength{\tabcolsep}{1.2mm}{
    \begin{tabular}{lccccc}
        \toprule
        Dataset & Metric & \makecell{DCGC - \\ w/o OA} & \makecell{DCGC - \\w/o O} & \makecell{DCGC -\\ w/o A} & DCGC \\
        \midrule
        \multirow{4}{*}{Cora} 
                            & ACC & $36.82_{\pm2.95}$ & $46.71_{\pm1.67}$ & $73.42_{\pm2.86}$ & $\bm{75.41}_{\pm\bm{1.69}}$ \\
                            
                            & NMI & $12.72_{\pm1.85}$ & $23.57_{\pm1.14}$ & $54.99_{\pm2.22}$ & $\bm{56.55}_{\pm\bm{1.20}}$ \\
       
                            & ARI &  $9.53_{\pm1.81}$ & $18.54_{\pm1.21}$ & $50.44_{\pm3.24}$ & $\bm{53.61}_{\pm\bm{1.94}}$ \\
        
                            & F1 & $35.22_{\pm3.02}$ & $43.51_{\pm1.63}$ & $69.66_{\pm4.23}$ & $\bm{72.15}_{\pm\bm{3.12}}$ \\
        \cmidrule{1-6}
        \multirow{4}{*}{ACM} & ACC  & $88.79_{\pm0.17}$ & $88.95_{\pm0.75}$ & $89.97_{\pm0.92}$ & $\bm{92.50}_{\pm\bm{0.32}}$\\
                            
                             & NMI & $63.08_{\pm1.6}$ & $63.42_{\pm1.27}$ & $ 66.10_{\pm0.51}$ & $\bm{72.98}_{\pm\bm{0.60}}$  \\
       
                           & ARI &  $69.39_{\pm2.31}$ & $69.81_{\pm1.78}$ & $72.48_{\pm1.10}$ & $\bm{78.87}_{\pm\bm{0.63}}$ \\
        
                            & F1 & $88.76_{\pm0.91}$ & $88.90_{\pm0.82}$ & $ 89.99_{\pm0.62}$ & $\bm{92.51}_{\pm\bm{0.23}}$ \\
        \cmidrule{1-6}
        \multirow{4}{*}{DBLP} & ACC  & $71.16_{\pm1.54}$ & $71.38_{\pm1.61}$ & $73.79_{\pm1.42}$ & $\bm{78.05}_{\pm\bm{0.56}}$ \\
                            
                             & NMI & $37.36_{\pm3.03}$ & $37.42_{\pm1.31}$ & $ 41.23_{\pm1.72}$ & $\bm{46.46}_{\pm\bm{1.22}}$  \\
       
                           & ARI &  $37.62_{\pm2.28}$ & $37.93_{\pm1.43}$ & $43.17_{\pm1.85}$ & $\bm{50.68}_{\pm\bm{0.94}}$ \\
        
                            & F1 & $70.88_{\pm1.53}$ & $71.07_{\pm1.34}$ & $ 73.51_{\pm1.55}$ & $\bm{77.51}_{\pm\bm{0.66}}$ \\
        \cmidrule{1-6}
        \multirow{4}{*}{Citeseer} & ACC  & $61.07_{\pm1.23}$ & $61.42_{\pm1.25}$ & $64.43_{\pm1.88}$ & $\bm{71.34}_{\pm\bm{0.30}}$ \\
                            
                             & NMI & $34.03_{\pm1.46}$ & $34.38_{\pm1.31}$ & $ 39.62_{\pm1.87}$ & $\bm{46.32}_{\pm\bm{0.48}}$ \\
       
                           & ARI &  $33.85_{\pm1.70}$ & $34.14_{\pm1.55}$ & $37.69_{\pm1.53}$ & $\bm{47.57}_{\pm\bm{0.55}}$ \\
        
                            & F1 & $57.98_{\pm1.44}$ & $58.15_{\pm1.48}$ & $ 58.34_{\pm1.49}$ & $\bm{65.81}_{\pm\bm{0.94}}$  \\
        \cmidrule{1-6}
        \multirow{4}{*}{Amazon} & ACC  & $52.73_{\pm1.71}$ & $54.42_{\pm1.19}$ & $75.64_{\pm1.20}$ & $\bm{80.17}_{\pm\bm{0.13}}$ \\
                            
                             & NMI & $36.82_{\pm2.44}$ & $39.03_{\pm1.74}$ & $65.85_{\pm2.09}$ & $\bm{71.77}_{\pm\bm{0.37}}$ \\
       
                           & ARI &  $25.46_{\pm2.44}$ & $27.33_{\pm1.34}$ & $57.19_{\pm2.64}$ & $\bm{66.70}_{\pm\bm{0.38}}$ \\
        
                            & F1 & $51.06_{\pm1.98}$ & $51.94_{\pm1.48}$ & $70.63_{\pm1.69}$ & $\bm{75.59}_{\pm\bm{0.34}}$ \\

        \cmidrule{1-6}
        \multirow{4}{*}{PubMed}
                        & ACC & $43.66_{\pm2.71}$ & $62.92_{\pm0.34}$ & $66.71_{\pm1.67}$ & $\bm{68.91}_{\pm\bm{0.43}}$ \\

                        & NMI & $ 3.37_{\pm1.67}$ & $27.69_{\pm1.41}$ & $30.16_{\pm2.90}$ & $\bm{32.53}_{\pm\bm{1.65}}$ \\

                        & ARI & $ 3.25_{\pm1.53}$ & $25.40_{\pm1.04}$ & $28.04_{\pm2.86}$ & $\bm{30.81}_{\pm\bm{0.85}}$ \\

                        & F1  & $43.33_{\pm2.92}$ & $63.51_{\pm0.62}$ & $66.16_{\pm1.83}$ & $\bm{68.74}_{\pm\bm{0.62}}$ \\
                            
    \bottomrule
    \end{tabular}}
    \label{tab:ablation_feature}
    \vspace{-4mm}
\end{table}

\begin{figure}
\centering
\subfloat[Cora]{
\includegraphics[width=0.3\textwidth]{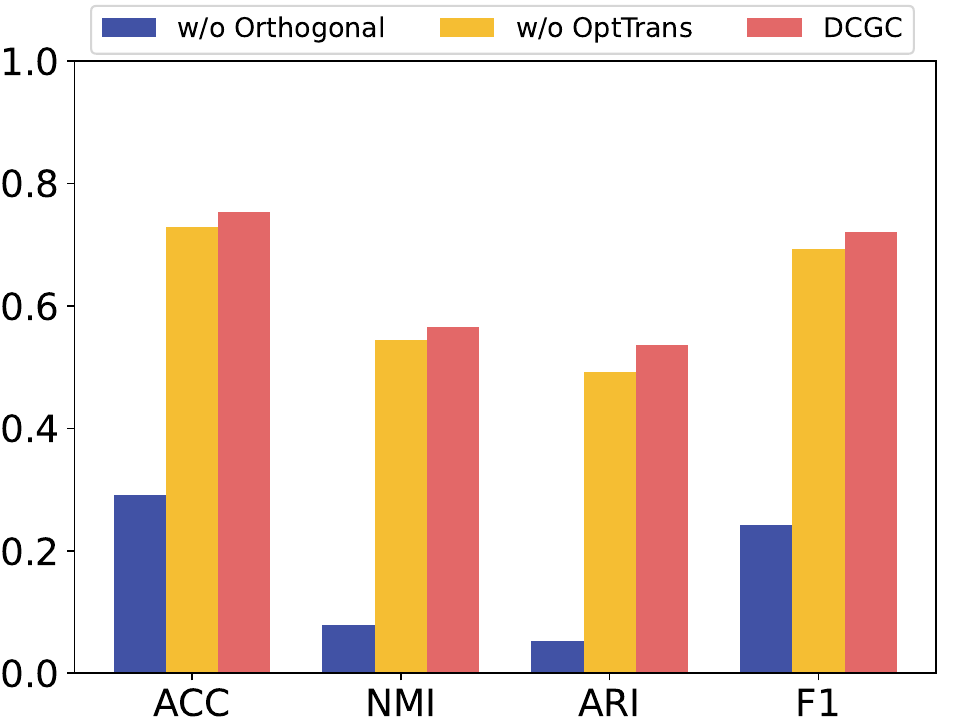}
}
\subfloat[ACM]{
\includegraphics[width=0.3\textwidth]{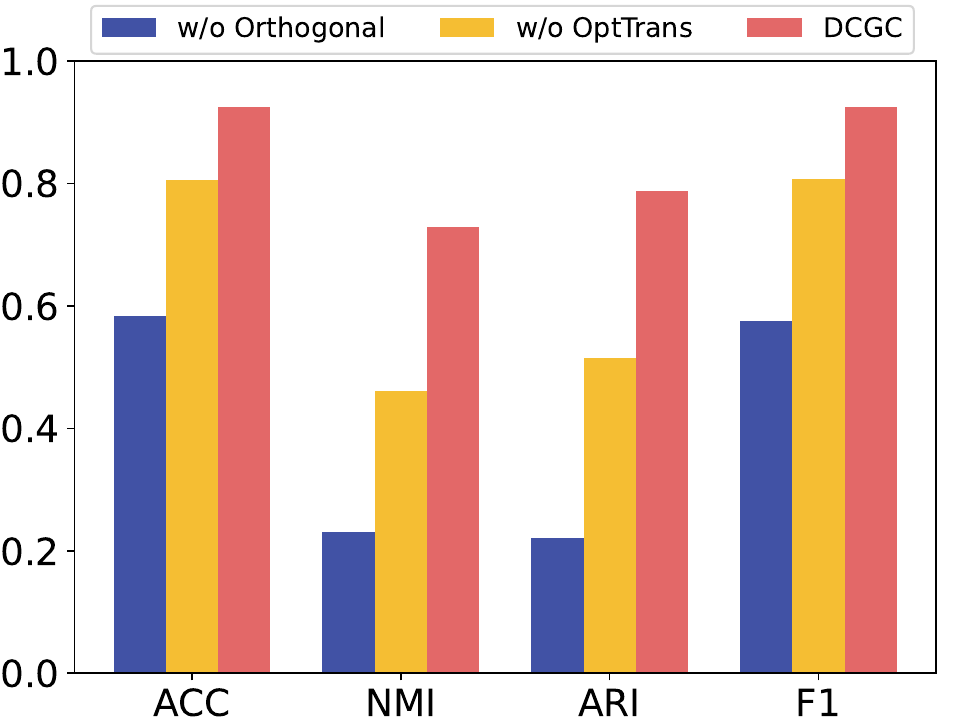}
}
\vspace{-3mm}
\subfloat[DBLP]{
\includegraphics[width=0.3\textwidth]{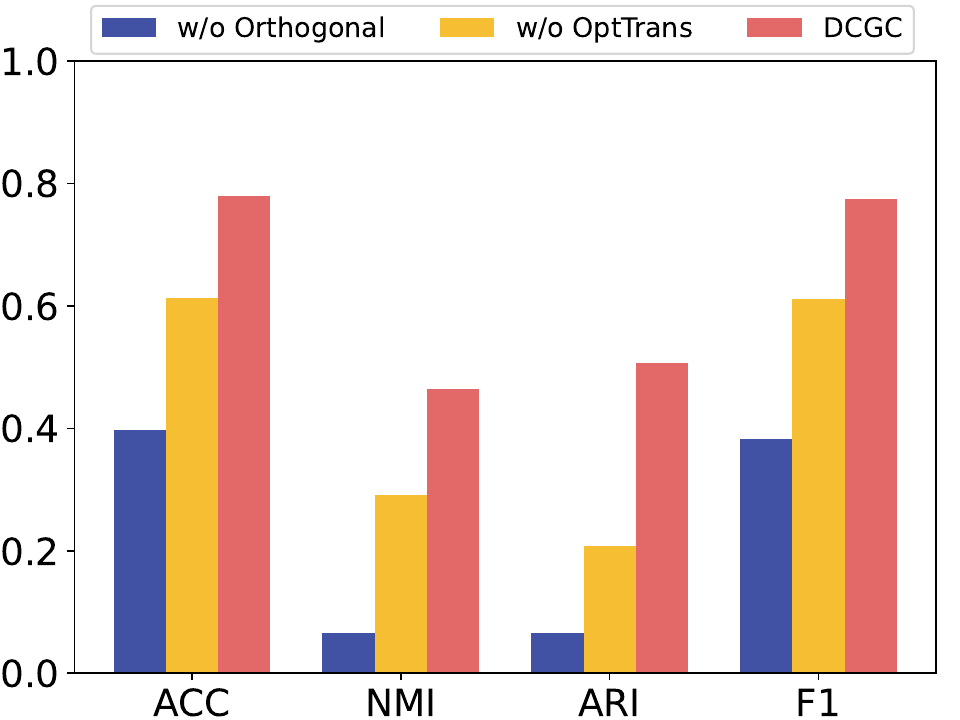}
}
\\
\subfloat[Citeseer]{
\includegraphics[width=0.3\textwidth]{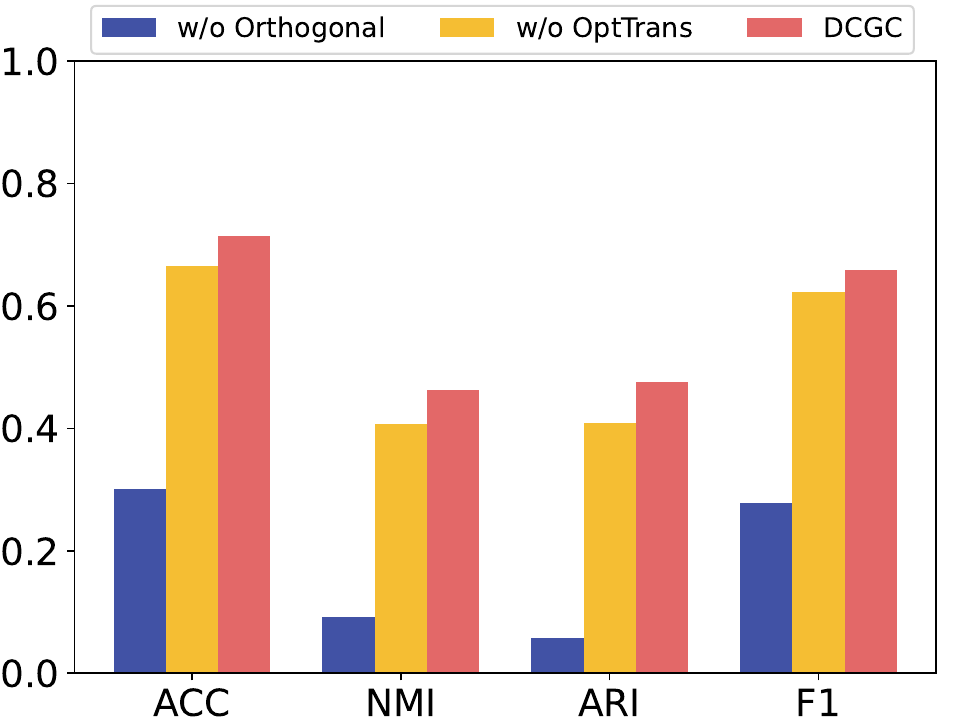}
}
\vspace{-3mm}
\subfloat[Amazon]{
\includegraphics[width=0.3\textwidth]{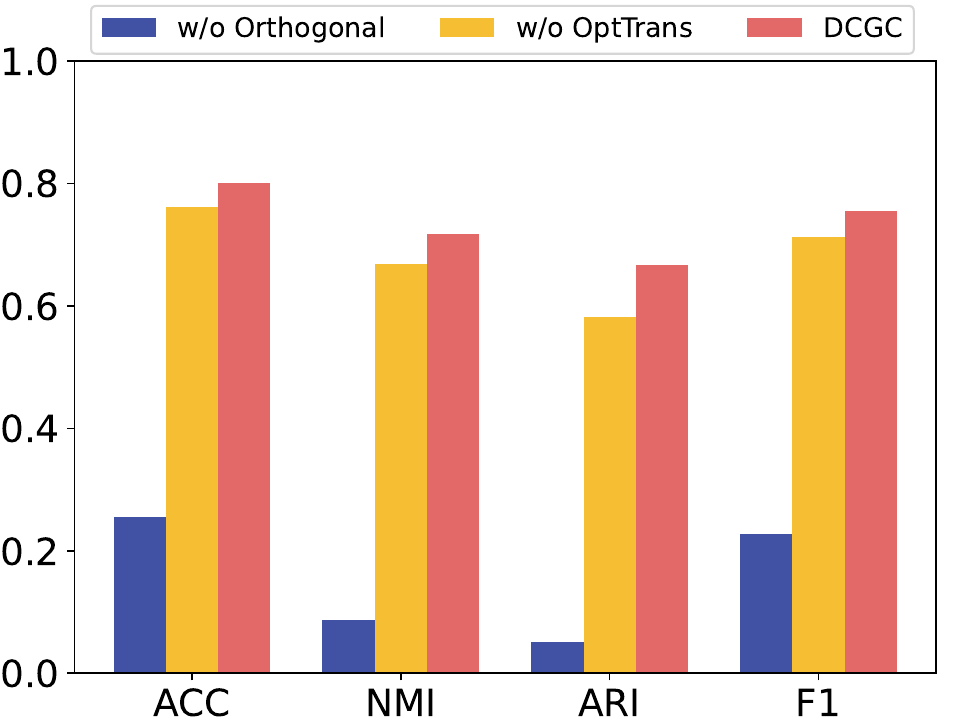}
}
\subfloat[PubMed]{
\includegraphics[width=0.3\textwidth]{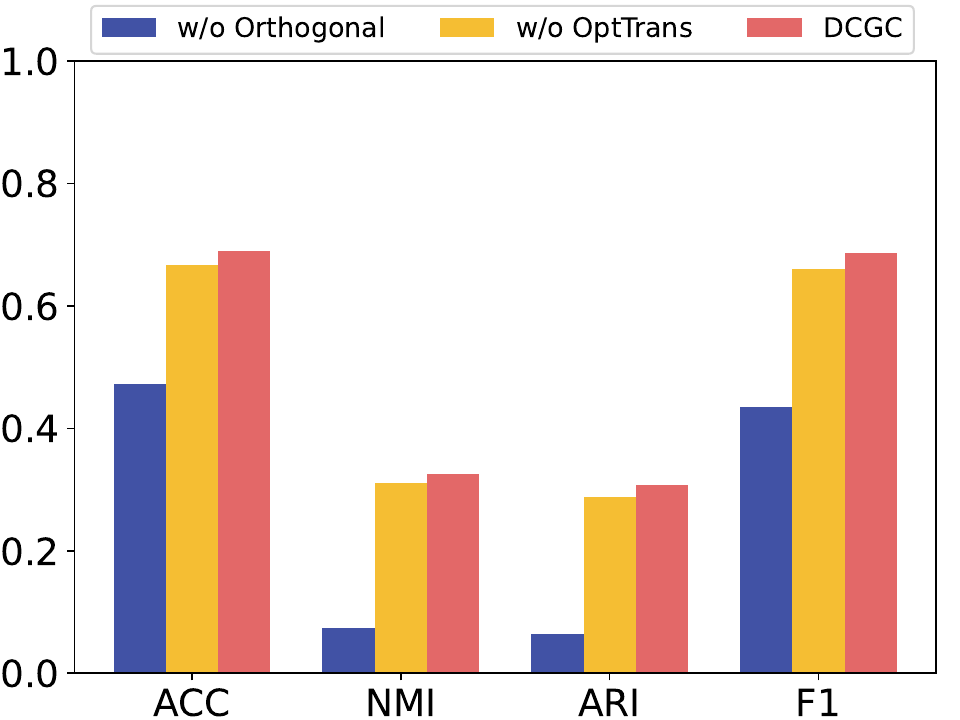}
}
\caption{Ablation comparisons of orthogonality regularization and optimal transport on six datasets.}
\label{fig:ablation}
\vspace{-4mm}
\end{figure}

\paragraph{Effect of orthogonality regularization.} From the perspective of minimizing normalized cut, the orthogonality of $\bm{H}$ is necessary. To further investigate the superiority of the proposed orthogonality regularization, we experimentally compare our method (i.e., DCGC in Figure \ref{fig:ablation} with its baseline, i.e., w/o Orthogonal in Figure \ref{fig:ablation}). Likewise, w/o Orthogonal is denoted that the baseline does not use the proposed orthogonality regularization. From the results in Figure \ref{fig:ablation}, we can see that DCGC outperforms w/o Orthogonal in terms of four metrics on six datasets. 
Such results indicate that the introduction of the orthogonality regularization in the cut-informed graph encoding module is necessary to make the learned node embeddings sufficiently different and thus avoid over-similarity and make better clustering results.

\paragraph{Effect of optimal transport.} To demonstrate our optimal transport based self-supervised graph clustering module is better than the traditional approaches (e.g., DEC~\cite{xie2016unsupervised} and SDCN~\cite{bo2020structural}) which adopt a clustering guided loss function to force the generated sample embeddings to have the minimum distortion against the pre-learned clustering center, we conduct ablation studies on six datasets with four metrics. In Figure \ref{fig:ablation}, w/o OptTrans is denoted that the baseline does not use the optimal transport based self-supervised graph clustering module but instead uses the traditional clustering guided loss function. From the results in Figure \ref{fig:ablation}, we can observe that DCGC outperforms w/o OptTrans in terms of four metrics on six datasets, which indicates that introducing optimal transport can balance the guidance of “proximity to the pre-learned cluster center”, thus avoid the degenerate solutions and achieve better results in graph clustering tasks.

\begin{figure}
\centering
\subfloat[Cora]{
\includegraphics[width=0.3\textwidth]{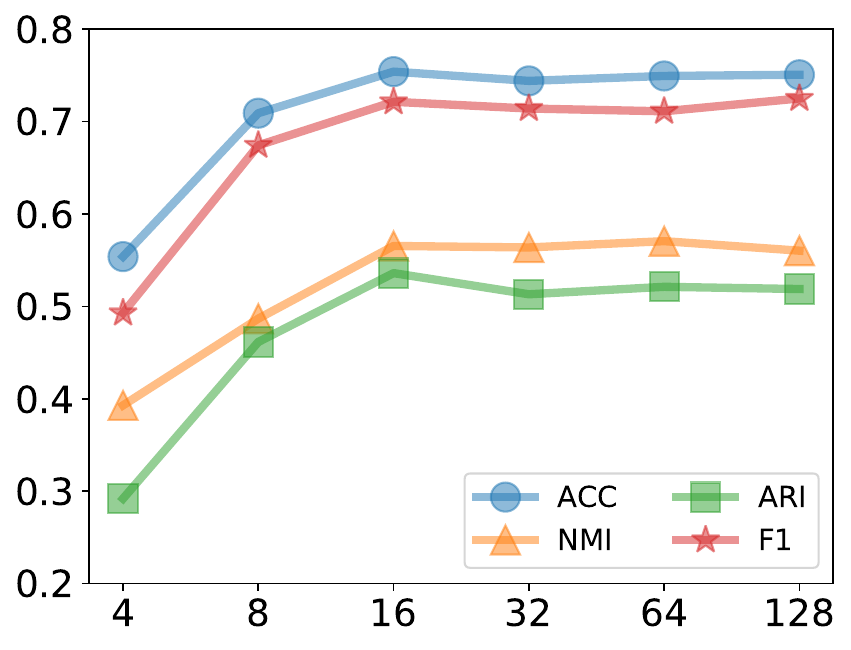}
}
\subfloat[ACM]{
\includegraphics[width=0.3\textwidth]{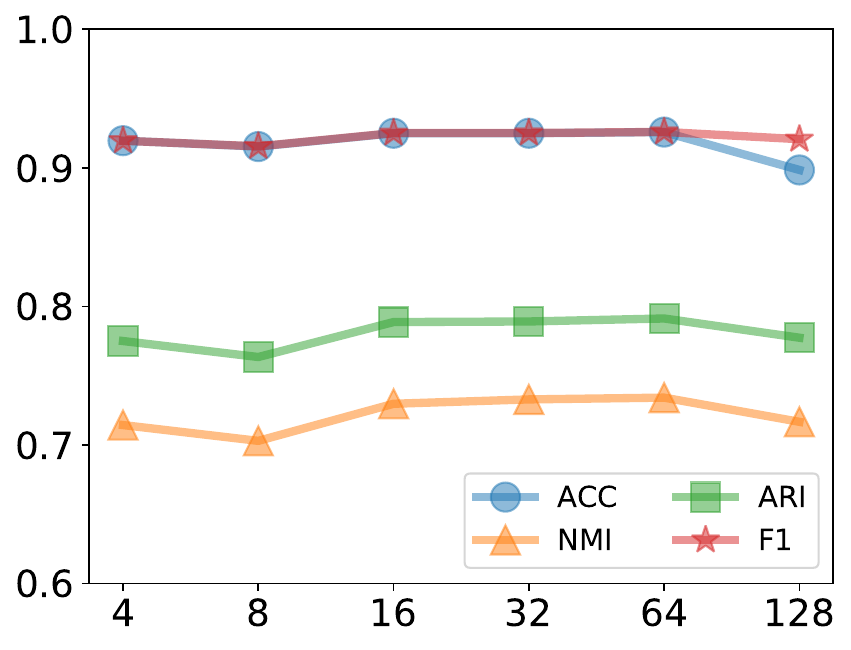}
}
\vspace{-0.3cm}
\subfloat[DBLP]{
\includegraphics[width=0.3\textwidth]{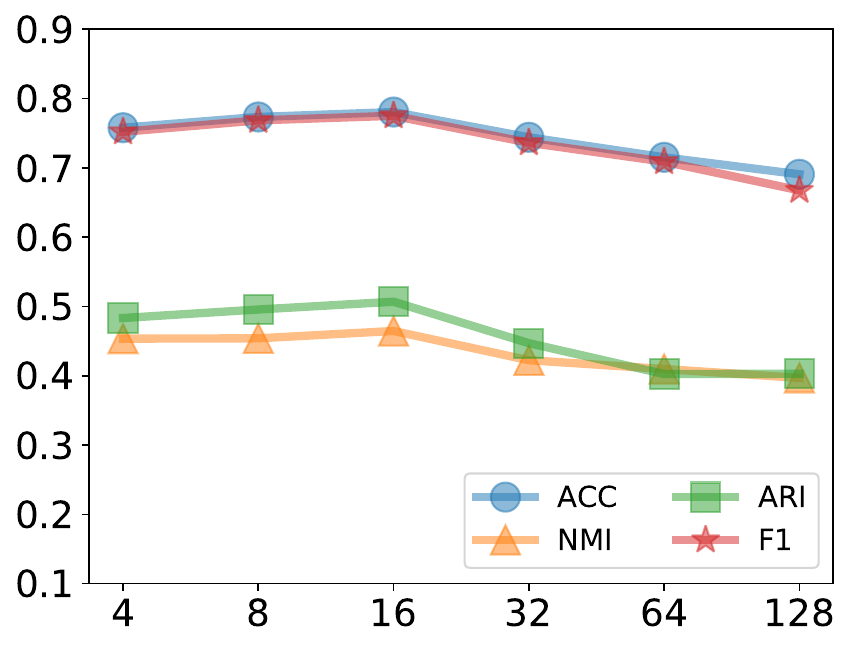}
}
\\
\subfloat[Citeseer]{
\includegraphics[width=0.3\textwidth]{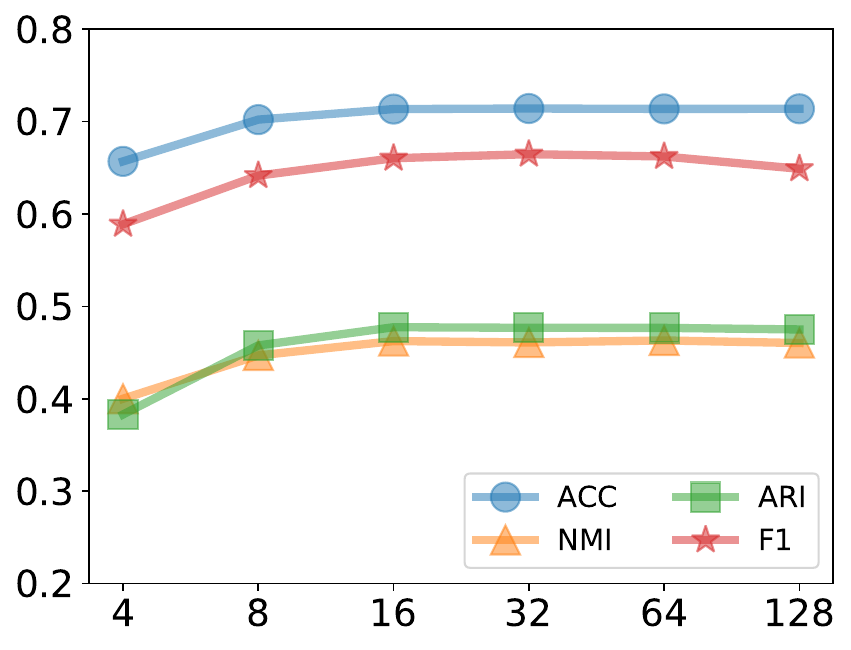}
}
\vspace{-0.3cm}
\subfloat[Amazon]{
\includegraphics[width=0.3\textwidth]{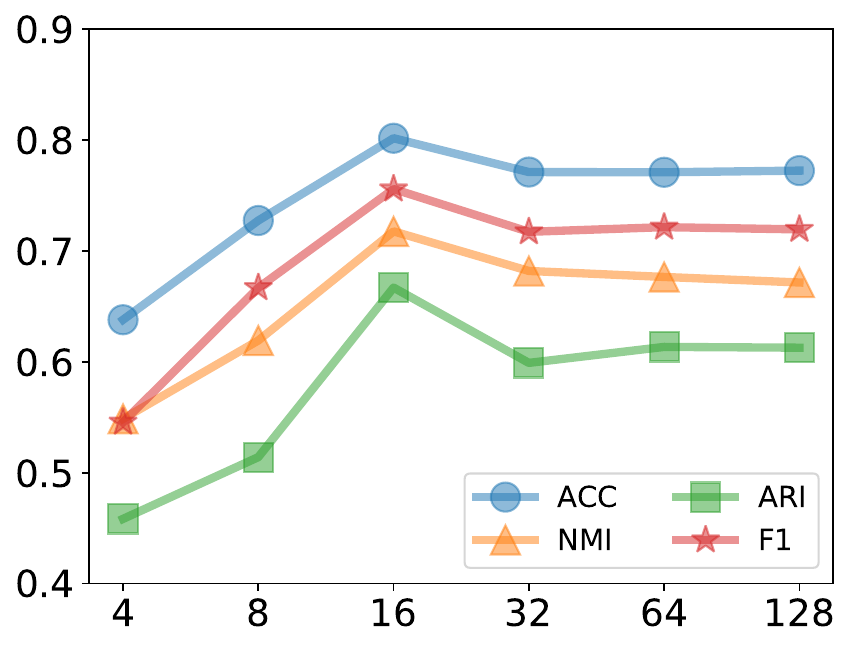}
}
\subfloat[PubMed]{
\includegraphics[width=0.3\textwidth]{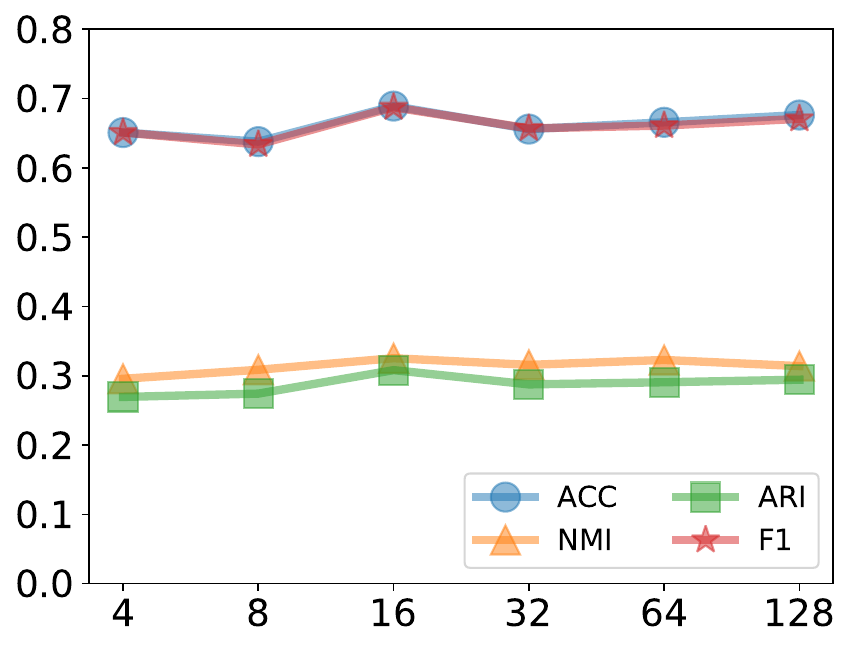}
}
\caption{Effect of dimension $d$ on four performance metrics.}
\vspace{-0.3cm}
\label{fig:dim}
\vspace{-4mm}
\end{figure}

\paragraph{Effect of embedding dimension.} The graph attributes are encoded as embedding, and the dimension of the embedding can influence computational time and costs, so it is reasonable to try to find a minimal dimension that does not affect performance. To discover the effect of dimension used for embedding on metric scores, we designed a series of experiments. We varied the dimension $d$ from 4 to 128 on the six datasets to see the clustering performance. According to the theoretical properties of normalized cut, the dimension of the embedding can be as small as the number of clusters. As shown in Figure~\ref{fig:dim}, the performance is still comparable even when $d = 4$ for ACM, DBLP, and PubMed, whose number of classes is 3, 4, and 3, respectively. As for Cora, there is a prominent rise in all 4 metrics when the dimension increases from 4 to 8, exceeding its number of classes, which is 7. Similar trends can also be observed on Citeseer and Amazon datasets. These observations are in accord with the theoretical properties of normalized cut. This implies that it is safe to use a small dimension as long as it is greater than the number of classes in the dataset.

\begin{figure*}[htbp!]
    \centering
    \includegraphics[width=1\textwidth]{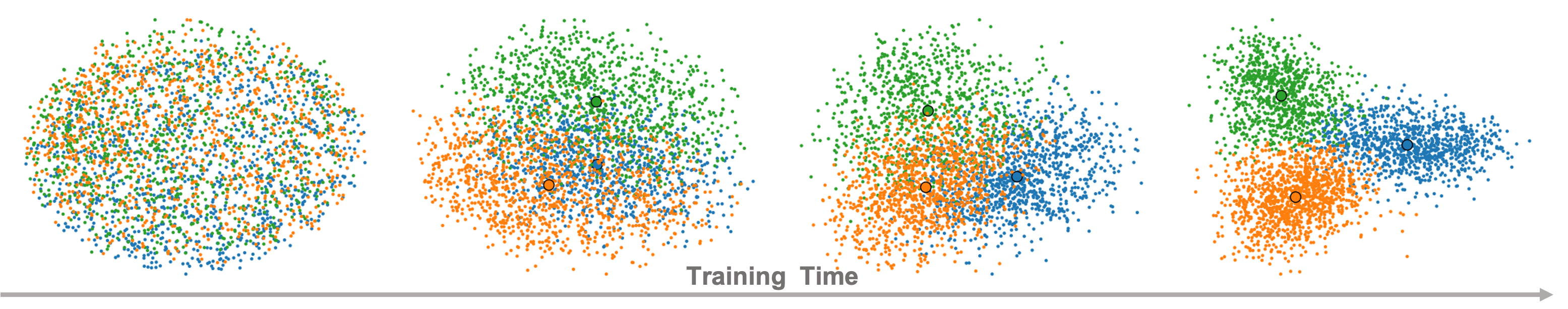}
    \caption{PCA visualization of learned embeddings when training on the ACM dataset. Black circles indicate the cluster centroids.}
    \label{fig:visual}
    \vspace{-4mm}
\end{figure*}

\noindent
\begin{figure}[htbp!]
  \centering
  \subfloat[Balance parameter \bm{$\alpha$}]{
    \includegraphics[width=0.24\textwidth]{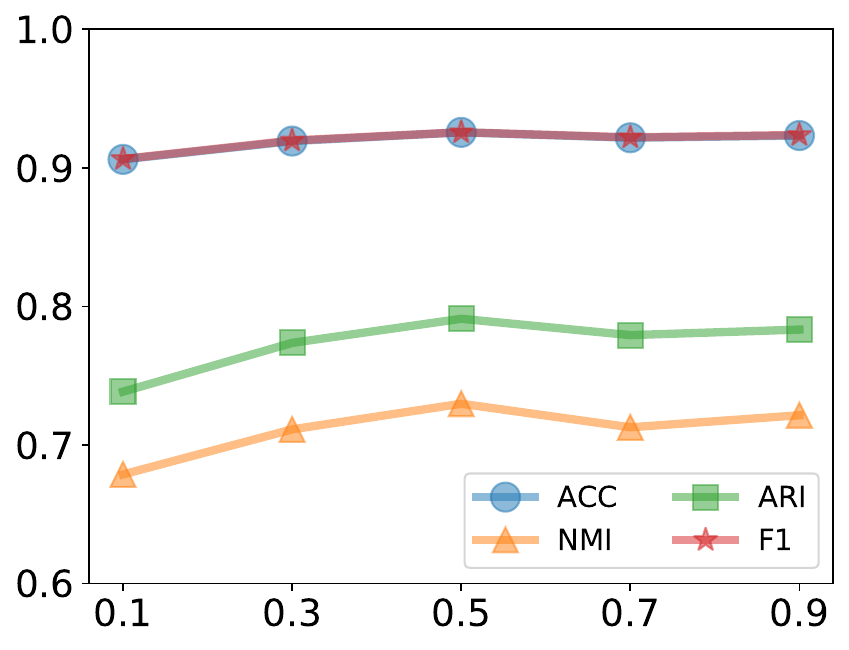}}
  \subfloat[Tuning parameter \bm{$\beta$}]{
    \includegraphics[width=0.24\textwidth]{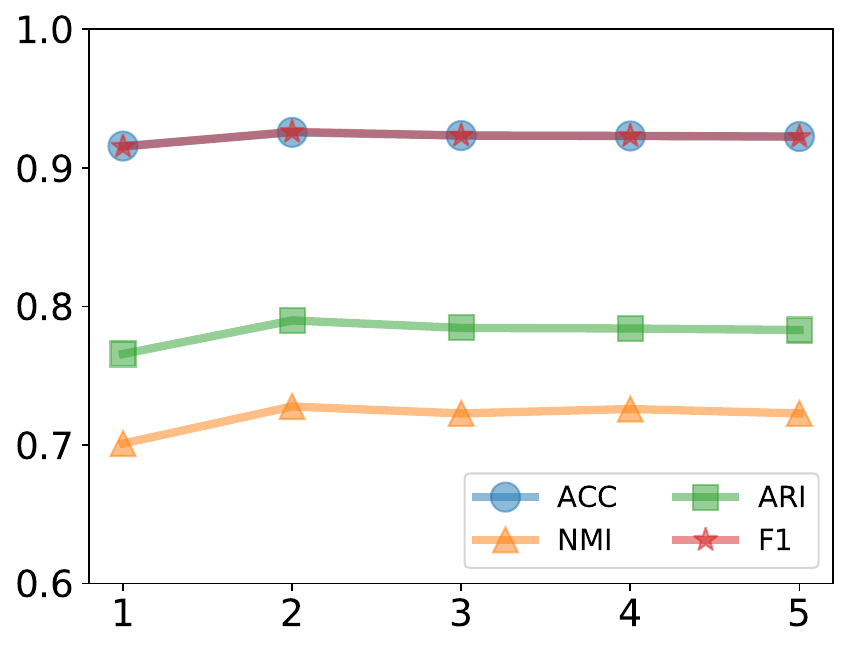}} 
  \subfloat[Tuning parameter \bm{$\gamma$}]{
    \includegraphics[width=0.24\textwidth]{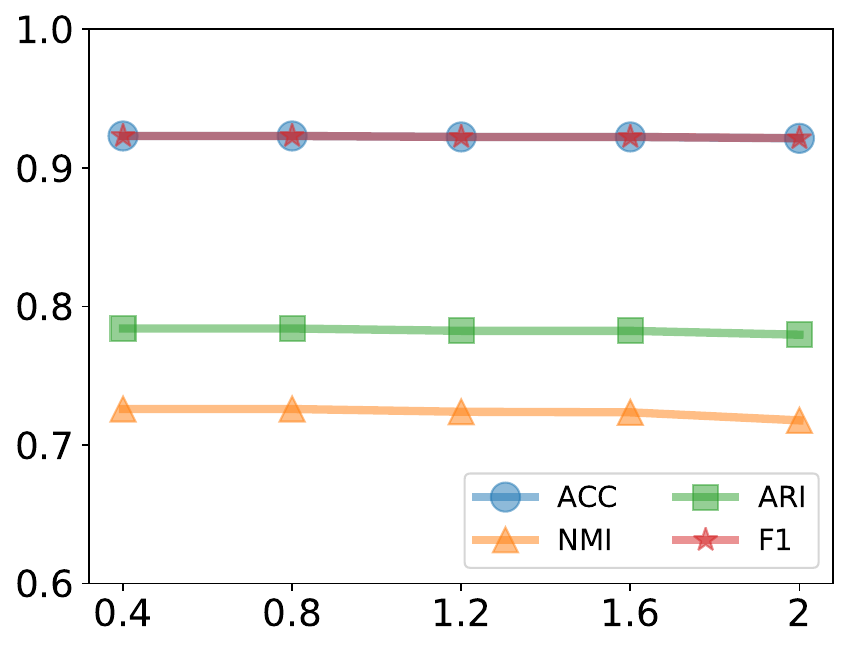}}
  \subfloat[Smoothness parameter \bm{$\lambda$}]{
    \includegraphics[width=0.24\textwidth]{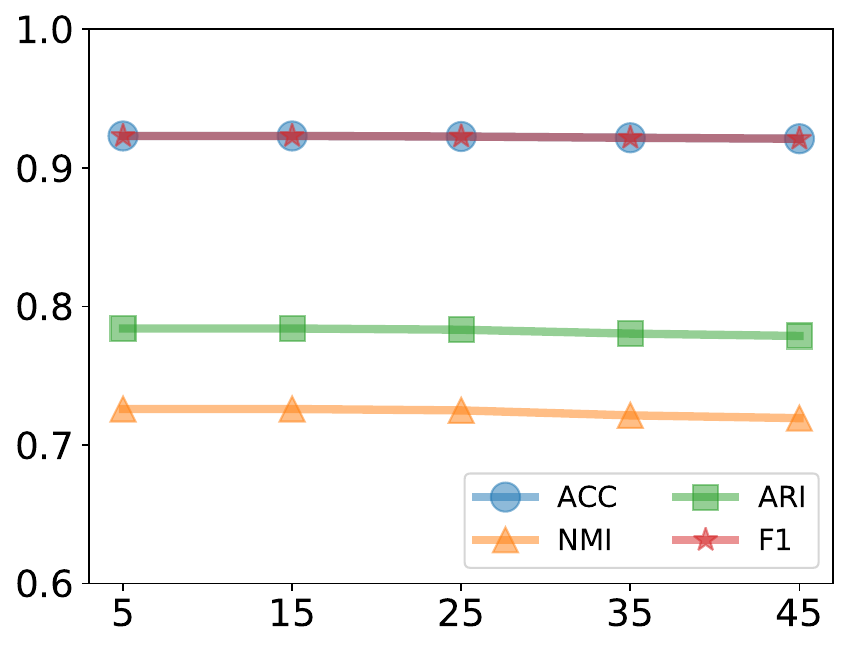}}
  \caption{Performance of DCGC on ACM benchmark datasets w.r.t. different hyper-parameters, including balance parameter $\alpha$, tuning parameter $\beta$, tuning parameter $\gamma$ and the smoothness parameter $\lambda$.}
  \label{fig:gamma}
  \vspace{-4mm}
\end{figure}

\vspace{-4mm}
\subsection{Visualization}
To show the superiority of the representation obtained by our proposed method, we use PCA to visualize the feature space.
The visualizations of the clusters on ACM are given in Figure \ref{fig:visual}. 
From left to right, they are the space of raw data, initialization embeddings, and learned embeddings (epoch 20 and epoch 100) of DCGC, respectively. 
We can see that the representations obtained by DCGC are discriminative, and each cluster is compact, indicating the clusters could be distinguished clearly in the feature space and revealing the intrinsic clustering structure among data.

\vspace{-4mm}
\subsection{Parameter Sensitivity}
As can be seen in Equation \ref{equ:loss_ge}, \ref{equ:opti_1} and \ref{equ:total_loss} that DCGC introduces four hyper-parameters including balance parameter $\alpha$, tuning parameter $\beta$, tuning parameter $\gamma$ and the smoothness parameter $\lambda$ to make a trade-off between the its modules. 
We conduct experiments to study the influence of these hyper-parameters on ACM datasets.
Figure \ref{fig:gamma}
illustrate the effect of $\alpha$, $\beta$, $\gamma$ and $\lambda$ varying from 0 to 1, 0.5 to 5, 0.2 to 2 and 5 to 50, respectively to study the performance variation of DCGC. 
From these figures, we can observe that the performance of the DCGC is stable in a wide range of hyper-parameter values.

%% file: 6_conclusion.tex
\section{Conclusion}
\textbf{Summary of methods.}
To address the issue that existing GNN-based deep graph clustering algorithms often encounter the
problem of representation collapse, wherein nodes belonging to distinct categories are frequently mapped to similar representations during the sample encoding process.
We propose a new and non-GNN-based graph clustering framework from the perspective of graph cut, which is more appropriate for the graph clustering task.
We show that the normalized cut minimization can be relaxed into solving $k$-smallest eigenvalues of graph Laplacian, where the eigenvectors is a continuous approximation of graph partition and can be viewed as cut-informed graph embeddings. 
Specifically, we propose a cut-informed graph embedding objective to minimize the joint normalized cut of the original graph and attribute graph. 
And we propose a self-supervised approach via optimal transport to learn better clustering assignments. 
With the above tailored designs, our framework is more suitable for the task of graph clustering, which can effectively alleviate the problem of representation collapse and achieve better performance.
Finally, we conduct extensive experiments on five challenging real-world graph datasets and show our approach can outperform the state-of-the-art deep graph clustering models. 

\noindent
\textbf{Limitations and future work.}
While DCGC has shown promising results on graph clustering tasks, there are several directions for future work. First, the current framework relies on computing the eigenvectors of the graph Laplacian, which can be computationally expensive for large graphs. Investigating more scalable approximations of the normalized cut objective could allow DCGC to handle larger datasets. Investigating more advanced fusion strategies beyond the weighted sum, especially for graphs with distinct multi-modal features, remains a valuable direction for future work. 
Second, the optimal transport clustering module provides a self-supervised clustering approach, but further exploring semi-supervised techniques may improve performance when limited labels are available. Additionally, the effectiveness of using cosine similarity for attribute graph construction on datasets with predominantly sparse or purely categorical features requires further empirical validation. 3
Finally, while DCGC focuses on clustering node representations, directly generating cluster assignments for new unseen nodes via inductive learning remains an open challenge. Developing inductive extensions to generalize DCGC to out-of-sample nodes would be useful for many practical graph clustering applications.
In summary, potential directions include scaling to larger graphs, incorporating limited supervision, handling dynamic graphs, and inductive clustering of new nodes. 
Furthermore, a systematic analysis of the method's robustness against varying levels of attribute noise and informativeness was not performed and constitutes an important area for future investigation. Addressing these limitations could further enhance DCGC's modeling capabilities and applicability to real-world graph clustering tasks. But the current model provides a solid foundation and demonstrates the promise of a non-GNN cut-informed approach.

\section{Acknowledgment}
This research was supported by National Natural Science Foundation of China (Grant No. 62406306) and the State Key Laboratory of Internet of Things for Smart City (University of Macau) No. SKL-IoTSC(UM)-2024-2026/ORP/GA02/2023.